\documentclass[journal]{IEEEtran}
\IEEEoverridecommandlockouts                              

\usepackage{amsmath}
\usepackage{amssymb}
\usepackage{amsfonts}
\usepackage{latexsym}
\usepackage{multicol}
\usepackage{lipsum}
\usepackage{accents}
\usepackage{wrapfig}
\usepackage{tabularx}
\usepackage{euscript}
\usepackage{color}
\usepackage{xcolor}
\usepackage{graphicx}
\usepackage{float}
\usepackage[caption = false]{subfig}	
\usepackage{epsfig}
\usepackage{hyperref}
\usepackage{setspace}
\usepackage{fancyhdr}
\usepackage{textcomp}
\usepackage{cite}
\usepackage{epstopdf}
\usepackage{esvect}
\DeclareGraphicsExtensions{.pdf,.eps,.jpg}

\begin{document}
\title{\LARGE \bf 	A Novel Potential Field Controller
					 for Use on Aerial Robots }

\author{Alexander C. Woods and Hung M. La, \textit{Senior Member, IEEE}%
\thanks{A. Woods and H. La are with the Advanced Robotics and Automation (ARA) Laboratory, Department of Computer Science and Engineering, University of Nevada, Reno, NV 89557-0312 USA. Corresponding author: H. La; E-mail: hla@unr.edu.}
}

\maketitle

\begin{abstract}
Unmanned Aerial Vehicles (UAV), commonly known as drones, have many potential uses in real world applications. Drones require advanced planning and navigation algorithms to enable them to safely move through and interact with the world around them. This paper presents an extended potential field controller (ePFC) which enables an aerial robot, or drone, to safely track a dynamic target location while simultaneously avoiding any obstacles in its path. The ePFC outperforms a traditional potential field controller (PFC) with smoother tracking paths and shorter settling times. The proposed ePFC's stability is evaluated by Lyapunov approach, and its performance is simulated in a Matlab environment. Finally, the controller is implemented on an experimental platform in a laboratory environment which demonstrates the effectiveness of the controller.
\end{abstract}

\section{Introduction}\label{S.intro}
\begin{figure}[t]
 \centering
  \includegraphics[width=0.75\columnwidth]{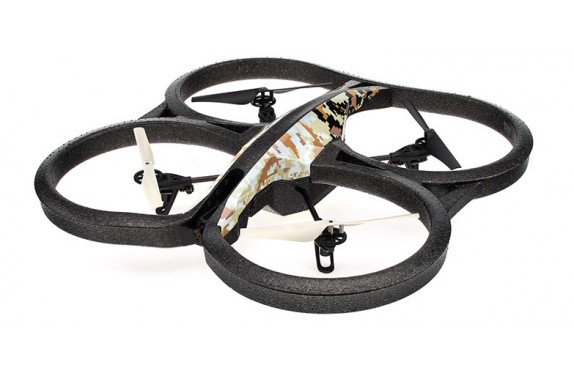}
  \caption{A low cost, commercially available quadcopter drone is used as the experimental platform for testing and demonstrating the effectiveness of the proposed control method.}
  \label{F.ARDrone}
\end{figure}

\IEEEPARstart{T}{his} paper focuses on dynamic target tracking and obstacle avoidance on a quadcopter drone such as the one shown in Fig.~\ref{F.ARDrone}. Recent advances in the field of unmanned autonomous systems (UAS) have drastically increased the potential uses of both unmanned ground vehicles (UGV) and unmanned aerial vehicles (UAV) \cite{Cui_SMCA_2016, Minaeian_SMCA_2016}. UAS can be utilized in situations which may be hazardous to human operators in ground vehicles or pilots in traditional aircraft, such as assisting wild land fire fighters~\cite{MartinezJ_2006,MerinoL_2006,CasbeerD_2005,SujitP_2007, YuanC_2015}, search and rescue operations in unsafe conditions or locations~\cite{PitreR_2012,BirkA_2011,TomicT_2012,GoodrichM_2008,ErdosD_2013}, and disaster relief efforts~\cite{QuaritschM_2010,MazaI_2011,BupeP_2015}. Additionally, UAS can be used in repetitive or tedious work where a human operator may lose focus such as infrastructure inspection~\cite{LaH_2013,MillsS_2011,LiZ_2010}, agricultural inspections~\cite{BlossR_2014,RoldanJ_2015}, and environmental sensing~\cite{Rossi_2014, La_SMCA_2015}. Specifically, quadcopter systems are desirable because they can perform very agile maneuvers which gives them an advantage over fixed wing platforms in confined environments.

Although the field of UAS has grown rapidly, it is still hindered by many problems which limit their use in real world applications. The challenge of localizing in GPS-denied environments has been approached by a multitude of research groups across the world, and there are several methods which have been  to address this. One of several promising on-board sensing methods is light detection and ranging (LIDAR). One group employed a reflexive algorithm in combination with a LIDAR sensor for simulating navigation through an unknown environment~\cite{ShaohuaM_2013}. Another group developed a multilevel simultaneous localization and mapping (SLAM) algorithm which utilized LIDAR as its primary sensing method~\cite{GrzonkaS_2012}. Other groups used LIDAR on autonomous vehicles for control and multi-floor navigation~\cite{Shen_2011,Sa_2012}.

Another major area of research for localization in GPS-denied environments is computer vision. One group used a single camera, looking at an object of known size to determine the drone's location~\cite{Kendall_2014}. Another group utilized a combination of LIDAR and a Microsoft Kinect sensor to explore an unknown environment~\cite{Shen_2012}. Several other groups successfully used unique variations of computer vision  methods as a means of localization~\cite{Gomez_2012,Meier_2012}, and it is proving to be a very promising method of operating in GPS-denied environments.

In addition to advanced sensing capabilities, UAS also require planning and navigation algorithms to safely move through and interact with the world around them. Trajectory generation for aerial robots has been accomplished through methods such as minimizing snap, the second derivative of acceleration~\cite{MellingerD_2011,MichaelN_2010}. Given keyframes consisting of a position in space coupled with a yaw angle, this method is able to generate very smooth, optimal trajectories. Other groups successfully applied methods utilizing Voronoi diagrams~\cite{Kendall_2014,Dogan_2003}, receding horizons in relatively unrestricted environments~\cite{Kuwata_2005}, high order parametric curves~\cite{NetoA_2013}, and 3D interpolation~\cite{AltmannA_2014}.

However, many platforms do not have the luxury of a very powerful processor and solving complex algorithms cannot practically be performed by an off-board computer. Therefore, the contribution of this paper is to propose an ePFC as a navigation method which is computationally inexpensive, can react quickly to the environment, and which can be deployed on-board any platform with adequate sensing capabilities. The ePFC expands on the basic capabilities of a traditional potential field controller, which operates only on $x$ and $y$ relative distances, and goes further to intelligently incorporate relative velocity as well. The developed controller's stability is evaluated, and its performance is both simulated and demonstrated experimentally. 

The remainder of this paper is organized as follows. Section \ref{S.model} presents a system model for the quadcopter system dynamics. Section~\ref{S.control} provides a brief background on potential field methods, discusses the design of the controller, and demonstrates the stability of the system using a Lyapunov approach. Section~\ref{S.simulation} presents simulation results of the controller implemented in a Matlab environment. Section~\ref{S.Experiment} discusses the experimental quadcopter platform, the testing environment, experimental results, and an evaluation of the controller's performance. Finally, section~\ref{S.conclusion} provides a brief conclusion, with recommendations for future work.

\section{System Model}\label{S.model}

This section presents the set of differential equations represent the quadcopter system dynamics. Developing a mathematical model of a system is a fundamental step in any controller design and develops a deeper understanding of the system in question. Once the mathematical system and initial controller design are complete, the combined system can be simulated and tested in an experimental setting. A brief background on Newtonian reference frames is provided, as well as the method used to transform between various reference frames and describe the motion of a rigid body. Finally, the equations of motion are derived using the Newton-Euler method.

\subsection{Reference Frames}\label{S.reference_frames}

\begin{figure}[htb]
  \centering
  \includegraphics[width=1\columnwidth]{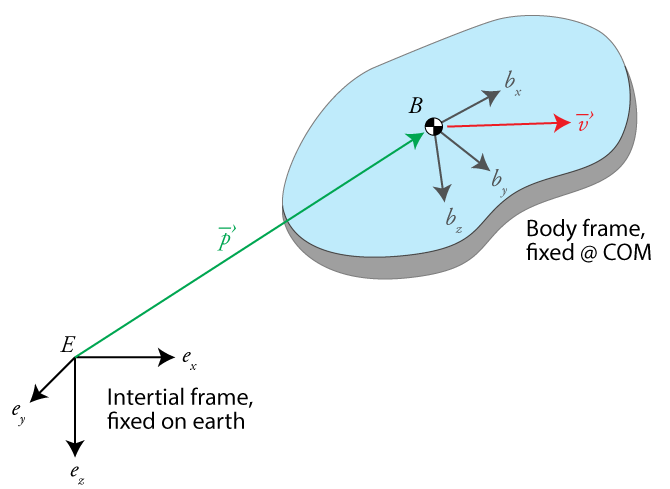}
  \caption{An inertial reference frame, $E$, is fixed on earth, and an accelerating body reference frame, $B$, is fixed at the rigid body's center of mass. The position and velocity of the rigid body are designated as $\vec{p}$ and $\vec{v}$ respectively. Euler rotation matrices may be used to transform between frames $E$ and $B$ for a given orientation.}
  \label{F.reference_frames} 
\end{figure}

To begin, it is important to introduce a set of reference frames which allow representation the position and orientation of a rigid body in space. These reference frames are defined by a linearly independent set of vectors which span the dimensions of the frame. The position and orientation of the rigid body at any instant in time is represented by a particle at its center of mass and is described relative to a Cartesian reference frame in Euclidean space, $E \in \mathbb{R}^{3}$, which is fixed on earth at a known location as shown in Fig.~\ref{F.reference_frames}. It is assumed that this reference frame is non-accelerating, and is therefore inertial. Additionally, the curvature of the earth is considered negligible for the scope of this work. The axes of $E \in \mathbb{R}^{3}$ are described by the set of orthogonal unit vectors $(\hat{e}_{x}, \hat{e}_{y}, \hat{e}_{z}) \in \mathbb{R}^{3}$, where $\hat{e}_{x}$ points north, $\hat{e}_{y}$ points east, and $\hat{e}_{z}$ points toward the center of the earth.

To simplify the derivation of the equations of motion, an additional reference frame, $B \in \mathbb{R}^{3}$, is defined which is attached to the particle at the center of mass of the rigid body. This reference frame is referred to as the body frame and is represented by the set of orthogonal unit vectors $(\hat{b}_{x}, \hat{b}_{y}, \hat{b}_{z}) \in \mathbb{R}^{3}$ where $\hat{b}_{x}$ points forward, $\hat{b}_{y}$ points right, and $\hat{b}_{z}$ points downward, perpendicular to the body. To describe the orientation of the body, name rotation about the $\hat{b}_{x}$ to be roll, $\vec{\phi}$, rotation about $\hat{b}_{y}$ to be pitch, $\vec{\theta}$, and rotation about $\hat{b}_{z}$ to be yaw, $\vec{\psi}$. It is important to note that the body frame is non-inertial and does experience acceleration.

\subsection{Euler Transformations}

In order to represent a vector in either reference frame, a transformation must be established between the two frames. Various methods exist for performing a transformation between frames, including Euler rotation matrices, quaternion transformations, and angle-axis representation. For the scope of this work, Euler rotation matrices are used, but their limitations are noted.

Three matrices fully describe the transformation between the body frame and the inertial frame: rotation about the $\hat{b}_{z}$ axis, rotation about the $\hat{b}_{x}$ axis, and rotation about the $\hat{b}_{y}$ axis. Rotation about the $\hat{b}_{z}$ axis (yaw) is a familiar example, common in two dimensional transformations, and can be described by

\begin{equation}\label{E.z_rot_matrix}
   \mathbf{R}_{\psi} =
   \begin{bmatrix}
    cos(\psi)    	& -sin(\psi) & 0 \\
    sin(\psi)   	&  sin(\psi) & 0 \\
    0           &  0     	& 1 
  \end{bmatrix}.
\end{equation}

Similarly, rotation about the $\hat{b}_{x}$ axis (roll) is described by
\begin{equation}\label{E.x_rot_matrix}
   \mathbf{R}_{\phi} =
   \begin{bmatrix}
    1       	&  0					& 0 \\
    0  			&  cos(\phi) 	& -sin(\phi) \\
    0           &  sin(\phi)  	& cos(\phi)
  \end{bmatrix}.
\end{equation}

Finally, rotation about the $\hat{b}_{y}$ axis (pitch) is described by
\begin{equation}\label{E.y_rot_matrix}
   \mathbf{R}_{\theta} =
   \begin{bmatrix}
    cos(\theta)	&  0	& sin(\theta) \\
    0  						&  1	&  \\
    -sin(\theta)	&  0   	& cos(\theta)
  \end{bmatrix}.
\end{equation}

While these three matrices fully described the transformation between the two coordinate frames, it is often more convenient to combine them into a single matrix for performing calculations. This final matrix is given by the product of (\ref{E.z_rot_matrix}), (\ref{E.x_rot_matrix}), and (\ref{E.y_rot_matrix}) which yields

\begin{equation}\label{E.full_rot_matrix}
   \mathbf{R}_{\phi,\theta,\psi} =
   \begin{bmatrix}
    C_{\psi}C_{\theta}	&  C_{\psi}S_{\phi}S_{\theta}-S_{\psi}C_{\phi}	& C_{\psi}C_{\phi}S_{\theta}+S_{\psi}S_{\phi} \\
    S_{\psi}C_{\theta}  &  S_{\psi}S_{\phi}S_{\theta}+C_{\psi}C_{\phi}	& S_{\psi}C_{\phi}S_{\theta}-C_{\psi}S_{\phi} \\
    -S_{\theta}			&  S_{\phi}C_{\theta}   						& C_{\phi}C_{\theta}
  \end{bmatrix}.
\end{equation}

where $S_{x} = sin(x)$ and $C_{x} = cos(x)$. A useful property of this rotation matrix is that $\mathbf{R}_{\phi,\theta,\psi}^{-1} = \mathbf{R}_{\phi,\theta,\psi}^{T}$ which can be used for transforming from the inertial frame to the body frame if needed. However, it should be noted that if $cos({\theta}) = 0$, a singularity occurs in $\mathbf{R}_{\phi,\theta,\psi}$ in which case one degree of freedom is lost. To address this shortcoming, other methods such as quaternion representations are often used when describing aerial robots. However, for the scope of this work, it is assumed that the quadcopter will not see large angles and therefore will not experience gimbal lock.

\subsection{Newton-Euler Equations}

A classic method of deriving the equations of motion in robotics is the use of the Newton-Euler equations. Combined, these equations fully describe both the translational and rotational dynamics of a rigid body.

Consider the reference frame $B$ which is attached to the particle at the robot's center of mass as discussed in Section~\ref{S.reference_frames}. The velocity of $B$ in $E$ is defined as

\begin{equation}
    \vec{v} = \frac{d\vec{p}}{dt},
\end{equation}

\noindent
where $\vec{p}$ is the position vector from the origin of $E$ to the origin of $B$ at the robot's center of mass. The translational momentum of $B$ is given by

\begin{equation}
    \vec{L} = m\vec{v},
\end{equation}

\noindent
where $m$ is the mass of the robot. Newton's second law states that the sum of the external forces acting upon an object equals the time rate of change of its translational momentum. Therefore, the translational dynamics of the robot can be described by

\begin{equation}\label{E.newton1}
\begin{aligned}
    \sum \vec{F} &= \frac{d}{dt}(\vec{L}),\\
    &= \frac{d}{dt}(m\vec{v}).
\end{aligned}
\end{equation}

For the scope of this paper, it is assumed that mass is time-invariant. Taking the derivative in the earth frame yields

\begin{equation}\label{E.newton2}
\begin{aligned}
    \sum \vec{F}_{E} &= m\frac{d\vec{v}}{dt_{E}},\\
    &= m\vec{a}_{E},
\end{aligned}
\end{equation}

\noindent
where $\vec{a}_{E}$ is linear acceleration in the earth frame. The derivative can also be taken in $B$ in order to represent the dynamics in the body frame which yields

\begin{equation}\label{E.newton3}
\begin{aligned}
    \sum \vec{F}_{B} &= m\left(\frac{d\vec{v}}{dt_{B}} + \vec{\omega}_{B} \times \vec{v}_{B}\right),\\
    &= m(\vec{a}_{B} + \vec{\omega}_{B} \times \vec{v}_{B}).
\end{aligned}
\end{equation}

Because $B$ is located at the center of mass of the body, $\vec{\omega}_{B}$ is zero and (\ref{E.newton3}) simplifies to be

\begin{equation}\label{E.newton4}
    \sum \vec{F}_{B} = m\vec{a}_{B}.
\end{equation}

While (\ref{E.newton4}) accurately describes the translational dynamics of the robot, the angular dynamics must still be addressed. The angular momentum of $B$ is defined as

\begin{equation}
    \vec{H} = I_{cm}\vec{\omega},
\end{equation}

\noindent
where $I_{cm}$ is the moment of inertia about the robot's center of mass. Euler's second law states that the sum of the torques acting upon an object equals the time rate of change its angular momentum. Therefore, the rotational dynamics are described using

\begin{equation}\label{E.euler1}
\begin{aligned}
    \sum \vec{\tau} &= \frac{d}{dt}(\vec{H}),\\
    &= \frac{d}{dt}(I_{cm}\vec{\omega}).
\end{aligned}
\end{equation}

Similar to mass, it is assumed that $I_{cm}$ is time-invariant. Therefore, the time derivative of $\vec{H}$, taking into account a rotating frame, is found as

\begin{equation}\label{E.euler2}
\begin{aligned}
    \sum \vec{\tau}_{B} &= I_{cm}\frac{d\vec{\omega}}{dt}_{B} + \vec{\omega}_{B} \times I_{cm}\vec{\omega}_{B},\\
    &= I_{cm}\vec{\alpha}_{B} + \vec{\omega}_{B} \times I_{cm}\vec{\omega}_{B},
\end{aligned}
\end{equation}

\noindent
where $\vec{\alpha}$ is angular acceleration.

It is common to combine (\ref{E.newton4}) and (\ref{E.euler2}) into matrix form for a more compact representation, given by

\begin{equation}\label{E.newton-euler}
   \begin{bmatrix}
    \sum \vec{F}_{B}\\
    \sum \vec{\tau}_{B}
  \end{bmatrix}
  = 
  \begin{bmatrix}
    mI_{3}    	& 0 \\
    0       	& I_{cm}
  \end{bmatrix}
  \begin{bmatrix}
    \vec{a}_{B}\\
    \vec{\alpha}_{B} 
  \end{bmatrix}
  +
  \begin{bmatrix}
    0\\
    \vec{\omega}_{B} \times I_{cm}\vec{\omega}_{B}
  \end{bmatrix},
\end{equation}

\noindent
where $I_{3}$ is a $3x3$ identity matrix. This is the classic form of the Newton-Euler equations for a system with $B$ oriented at the system's center of mass, and is the basis for determining the equations of motion specific to the quadcopter platform.

\subsection{Quadcopter Dynamics}

\begin{figure}[htb]
  \centering
  \includegraphics[width=0.75\columnwidth]{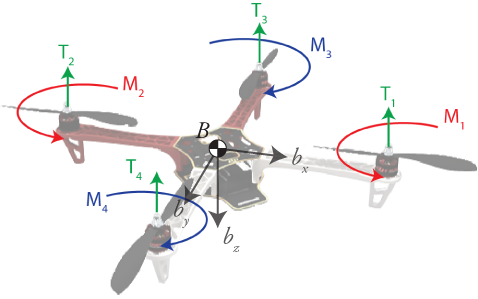}
  \caption{A quadcopter is symmetric about both the $\hat{b}_{x}$ and $\hat{b}_{y}$ axis. Each motor produces a force, $\vec{T}_{i}$ and moment, $\vec{M}_{i}$, which are functions of the motor's angular velocity, $\Omega_{i}$.}
  \label{F.multirotor2}
\end{figure}

The quadcopter platform shown in Fig~\ref{F.multirotor2} is assumed to be symmetric about the $\hat{b}_{x}$ and $\hat{b}_{y}$ axis. Thus the inertia matrix in the body frame is given by

\begin{equation}
   I_{cm} = 
  \begin{bmatrix}
    I_{xx}    & 0 		  & 0 \\
    0       	& I_{yy}	& 0 \\
    0			    & 0			  & I_{zz}
  \end{bmatrix},
\end{equation}

\noindent
where $I_{xx}$ and $I_{yy}$ are equal due to symmetry.

Each of the four motors on the quadcopter has a force associated with it, given by

\begin{equation}
	\vec{T}_{i} = -k_{T}\Omega_{i}^{2}\hat{b_{z}},
\end{equation}

\noindent
where $\vec{T_{i}}$ is the force (or thrust) provided by the $i^{th}$ motor, $\Omega_{i}$ is the angular velocity of the motor, and $k_{T}$ is a constant which is a function of the specific motor and propeller used. For the scope of this paper, it is assumed that the density of air remains constant and that roll and pitch are small angles. Using this, sum of the forces is found to be

\begin{equation}\label{E.sum_forces}
	\sum\vec{F} = 
	\begin{bmatrix}
    0 & 0 & 0 \\
    0 & 0 & 0 \\
    0 & 0 & mg - k_{T}\sum\limits_{i=1}^{4}\Omega_{i}^{2}
  \end{bmatrix}
  \begin{bmatrix}
    \hat{b_{x}} \\
    \hat{b_{y}} \\
    \hat{b_{z}}
  \end{bmatrix}.
\end{equation}

\begin{equation}\label{E.sum_torques}
\resizebox{.9\columnwidth}{!}{$\displaystyle{
	\sum\vec{\tau} = 
	\begin{bmatrix}
	k_{T}(\Omega_{3}^{2}-\Omega_{4}^{2})l & 0 & 0 \\ 
	0 & k_{T}(\Omega_{1}^{2}-\Omega_{2}^{2})l & 0 \\
	0 & 0 & -k_{M}(\Omega_{1}^{2}+\Omega_{2}^{2}-\Omega_{3}^{2}-\Omega_{4}^{2})
	\end{bmatrix}
	\begin{bmatrix}
    \hat{b_{x}} \\
    \hat{b_{y}} \\
    \hat{b_{z}}
  \end{bmatrix},
}$}
\end{equation}

\begin{equation}\label{E.eom}
\resizebox{.9\columnwidth}{!}{$\displaystyle{
\begin{aligned}
	\begin{bmatrix}
    \vec{a}_{x} \\
    \vec{a}_{y} \\
    \vec{a}_{z} \\
    \vec{\alpha}_{x} \\
    \vec{\alpha}_{y} \\
    \vec{\alpha}_{z} \\
  \end{bmatrix}
  &=
  \begin{bmatrix}
    \frac{1}{m} & 0 & 0 & 0 & 0 & 0 \\
    0 & \frac{1}{m} & 0 & 0 & 0 & 0 \\
    0 & 0 & \frac{1}{m} & 0 & 0 & 0 \\
    0 & 0 & 0 & \frac{1}{I_{xx}} & 0 & 0 \\
	0 & 0 & 0 & 0 & \frac{1}{I_{yy}} & 0 \\
	0 & 0 & 0 & 0 & 0 & \frac{1}{I_{zz}}
  \end{bmatrix}
  \left(
  \begin{bmatrix}
    0 \\
    0 \\
    mg - k_{T}\sum\limits_{i=1}^{4}\Omega_{i}^{2} \\
    k_{T}(\Omega_{3}^{2}-\Omega_{4}^{2})l \\ 
	k_{T}(\Omega_{1}^{2}-\Omega_{2}^{2})l \\
	-k_{M}(\Omega_{1}^{2}+\Omega_{2}^{2}-\Omega_{3}^{2}-\Omega_{4}^{2})
  \end{bmatrix}
  -
  \begin{bmatrix}
    0 \\
    0 \\
    0 \\
    \dot{\theta}\dot{\psi}(I_{zz} - I_{yy}) \\ 
	\dot{\phi}\dot{\psi}(I_{xx} - I_{zz}) \\
	\dot{\theta}\dot{\phi}(I_{yy} - I_{xx})
  \end{bmatrix}
  \right)\\
  &=
  \begin{bmatrix}
    0 \\
    0 \\
    g - \frac{k_{T}}{m}\sum\limits_{i=1}^{4}\Omega_{i}^{2} \\
    \frac{1}{I_{xx}}\left(k_{T}(\Omega_{3}^{2}-\Omega_{4}^{2})l - \dot{\theta}\dot{\psi}(I_{zz} - I_{yy})\right) \\ 
	\frac{1}{I_{yy}}\left(k_{T}(\Omega_{1}^{2}-\Omega_{2}^{2})l - \dot{\phi}\dot{\psi}(I_{xx} - I_{zz})\right) \\
	\frac{1}{I_{zz}}\left(-k_{M}(\Omega_{1}^{2}+\Omega_{2}^{2}-\Omega_{3}^{2}-\Omega_{4}^{2}) - \dot{\theta}\dot{\phi}(I_{yy} - I_{xx})\right)
  \end{bmatrix}.
\end{aligned}
}$}
\end{equation}

Similar to thrust, each motor also has an associated moment, given by

\begin{equation}
	\vec{M}_{i} = -k_{M}\Omega_{i}^{2}\hat{b_{z}},
\end{equation}

where $\vec{M_{i}}$ is the moment generated by the $i^{th}$ motor, and $k_{M}$ is a constant which is again a function of the specific motor and propeller used. In addition to the moments generated by the motors themselves, there are torque contributions from the moment arms produced by the motor's forces. The sum of the torques can be found in (\ref{E.sum_torques}), where $l$ is the length of the quadcopter's arms.

From (\ref{E.sum_forces}) and (\ref{E.sum_torques}), it is apparent that the linear translational dynamics are closely coupled with the rotational dynamics. This is expected because a quadcopter is an underactuated system, having only four actuators and six degrees of freedom.

Using the result of (\ref{E.sum_forces}) and (\ref{E.sum_torques}) in the Newton-Euler equations, the dynamics of the system in the body frame, $B$, are described by (\ref{E.eom}), assuming small roll and pitch angles. These equations fully describe the motion of the quadcopter platform in the body coordinates, and can be used for simulating the dynamics of the platform to evaluate controller performance.

\section{Controller Design and Stability}\label{S.control}
Because of their simplicity and elegance, potential field controllers (PFCs) are often used for navigation of ground robots~\cite{LaH_2013a,LaH_2013b,LaH_2012}. Potential fields are aptly named, because they use attractive and repulsive potential equations to draw the drone toward a goal (attractive potential) or push it away from an obstacle (repulsive potential). For example, imagine a stretched spring which connects a drone and a target. Naturally, the spring draws the drone to the target location.

\begin{figure}[htb]
 \centering
  \includegraphics[width=1\columnwidth]{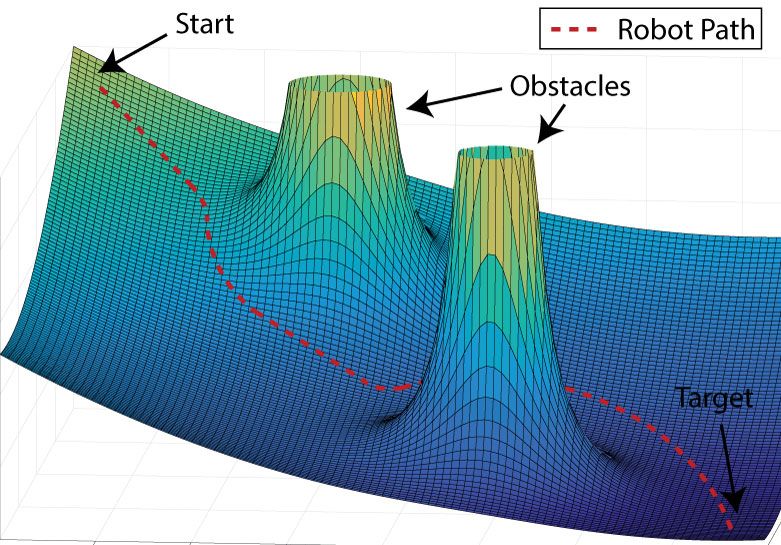}
  \caption{An example of a traditional potential field which can be used for navigating toward a target while avoiding multiple obstacles.}
  \label{F.potential}
\end{figure}

Conveniently, potential fields for both attractive and repulsive forces can be summed together, to produce a field such as the one shown in Fig.~\ref{F.potential}. This figure illustrates how a robot can navigate toward a target location while simultaneously avoiding obstacles in its path.

\subsection{Controller Design}
Denote $\vec{p}_d = [x_d, y_d, z_d]^T$ and $\vec{p}_t = [x_t, y_t, z_t]^T$ as the position vector of drone and target, respectively. The relative distance vector between the drone and the target is then

\begin{equation}
\begin{aligned}
\vec{p}_{dt} &= [x_{dt}, y_{dt}, z_{dt}]^T ,\\
&= [x_d, y_d, z_d]^T -  [x_t, y_t, z_t]^T.
\end{aligned}
\end{equation}

Traditionally, potential forces work in the $x$, $y$, and $z$ spatial dimensions, and are defined by a quadratic function given by

\begin{equation}\label{E.attractive}
    U_{att^{1}}(\vec{p}_{d},\vec{p}_{t}) = \frac{1}{2}\lambda_{1}\|\vec{p}_{dt}\|^2,
\end{equation}

\noindent
where $\lambda_{1}$ is positive scale factor, and $\|\vec{p}_{dt}\|$ is the magnitude of the relative distance between the drone and the target, which is given by
\begin{equation}\label{E.rel_distance}
    \|\vec{p}_{dt}\| = \sqrt{(x_{dt})^{2}+(y_{dt})^{2}+(z_{dt})^{2}}.
\end{equation}

As shown in Fig.~\ref{F.potential}, the target location is always a minimum, or basin, of the overall potential field. Therefore, in order to achieve the target location, the UAS should always move ``downhill.'' The direction and magnitude of the desired movement can be computed by finding the negative gradient of the potential field, given by

\begin{equation}\label{E.velocity_att}
\begin{aligned}
    \vec{v}_{d}^{att^{1}}(\vec{p}_{d},\vec{p}_{t}) &= -\nabla U_{att^{1}}(\vec{p}_{d},\vec{p}_{t}),\\
    &= -\frac{\partial U_{att^{1}}}{\partial x}\hat{i} - \frac{\partial U_{att^{1}}}{\partial y}\hat{j} - \frac{\partial U_{att^{1}}}{\partial z}\hat{k},\\
    &= -\nabla \Big(\frac{1}{2}\lambda_{1} \|\vec{p}_{dt}\|^{2}\Big),\\
    &= -\frac{1}{2}\lambda_{1}\nabla \|\vec{p}_{dt}\|^{2},\\
    &= -\frac{1}{2}\lambda_{1}\nabla (\vec{x}_{dt}^{2}+\vec{y}_{dt}^{2}+\vec{z}_{dt}^{2}),\\
    &= -\lambda_{1} (\vec{x}_{dt}+\vec{y}_{dt}+\vec{z}_{dt}),\\
    &= -\lambda_{1} (\vec{p}_{d} - \vec{p}_{t}),
\end{aligned}
\end{equation}

\noindent
where $\vec{v}_{d}^{att^{1}}$ is the desired velocity due to the attractive position potential.

This is the classic form of a simple attractive potential field controller. However, this does not yet take into account obstacles or other sources of repulsive potential. The repulsive potential is proportional to the inverse square of the distance between the drone and the obstacle and is given by

\begin{equation}\label{E.repulsive}
    U_{rep^{1}}(\vec{p}_{d},\vec{p}_{o}) = \frac{1}{2}\eta_{1}\frac{1}{\|\vec{p}_{do}\|^2},
\end{equation}

\noindent
where $\eta_{1}$ is positive scale factor, and $\|\vec{p}_{do}\|$ is the magnitude of the relative distance between the drone and the obstacle.

To find the desired velocity, again take the gradient of the potential field which yields
\begin{equation}\label{E.velocity_rep}
\begin{aligned}
    \vec{v}_{d}^{rep^{1}}(\vec{p}_{d},\vec{p}_{o}) &= -\nabla U_{rep^{1}}(\vec{p}_{d},\vec{p}_{o}),\\
    &= -\frac{\partial U_{rep^{1}}}{\partial x}\hat{i} - \frac{\partial U_{rep^{1}}}{\partial y}\hat{j} - \frac{\partial U_{rep^{1}}}{\partial z}\hat{k},\\
    &= -\nabla \Big(\frac{1}{2}\eta_{1}\frac{1}{\|\vec{p}_{do}\|^2}\Big),\\
    &= -\frac{1}{2}\eta_{1}\nabla \frac{1}{\|\vec{p}_{do}\|^2},\\
    &= -\frac{1}{2}\eta_{1}\nabla \frac{1}{\vec{x}_{do}^{2}+\vec{y}_{do}^{2}+\vec{z}_{do}^{2}},\\
    &= \eta_{1} \frac{\vec{x}_{do}+\vec{y}_{do}+\vec{z}_{do}}{(\vec{x}_{do}^2+\vec{y}_{do}^2+\vec{z}_{do}^2)^2},\\
    &= \eta_{1} \frac{\vec{p}_{do}}{\|\vec{p}_{do}\|^4},
\end{aligned}
\end{equation}

\noindent
where $\vec{v}_{d}^{rep^{1}}$ is the desired velocity due to the repulsive position potential. 

\begin{figure}[htb]
 \centering
  \includegraphics[width=1\columnwidth]{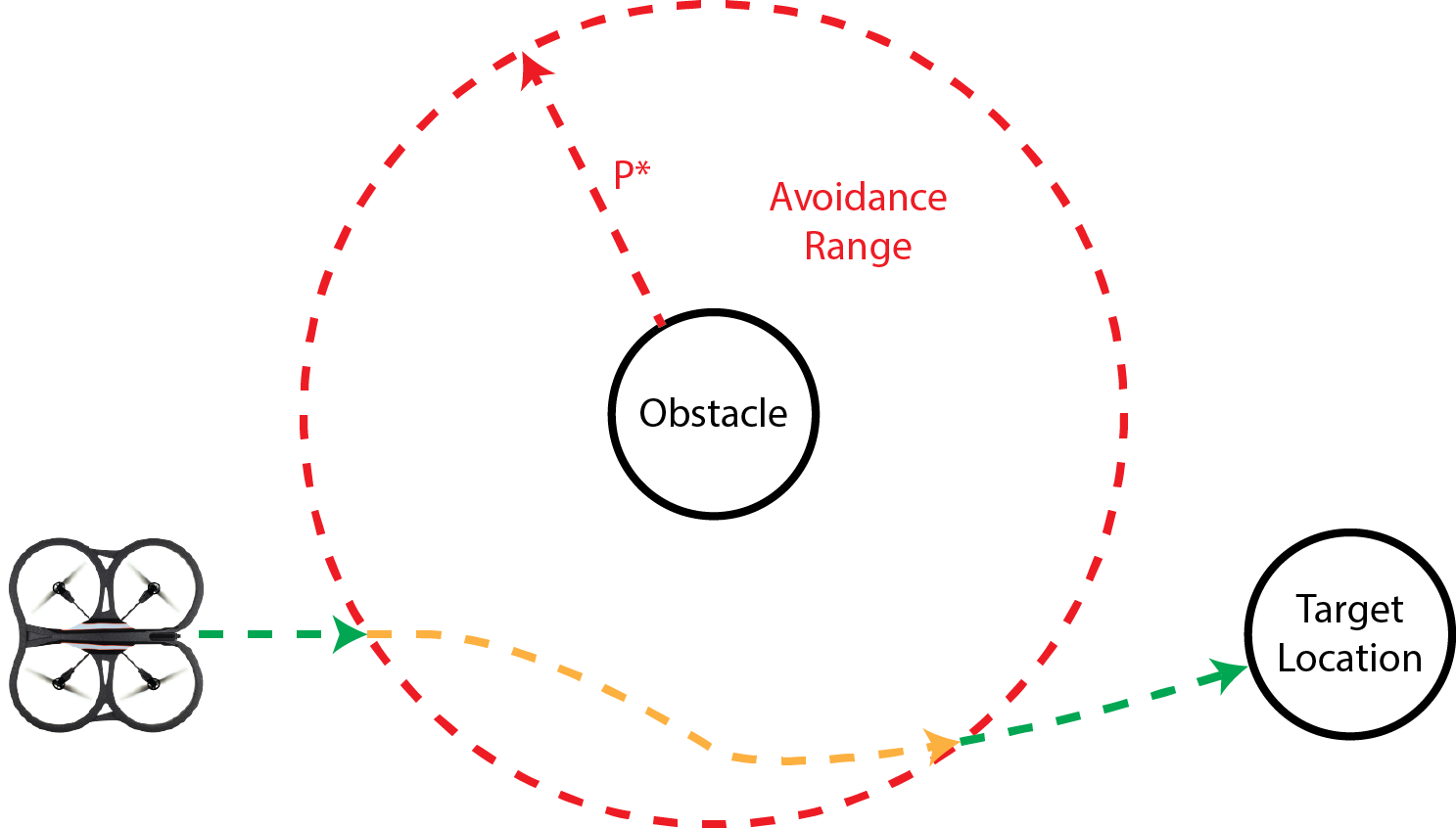}
  \caption{The repulsive force due to an obstacle should be zero when the drone is sufficiently far from it. In this case, the limit is $P^{*}$. While outside of this limit, the drone is unaffected by the obstacle's repulsive potential. However, as soon as the relative distance becomes less than $P^{*}$, the drone takes avoidance action.}
  \label{F.repulsive_limit}
\end{figure}

It is important to note that the repulsive potential is designed to have no effect when the drone is more than a set distance away from the obstacle, $P^{*}$, as shown in Fig.~\ref{F.repulsive_limit}.

Therefore, the velocity due to repulsive potentials becomes

\begin{equation}\label{E.velocity_rep_limit}
    \vec{v}_{d}^{rep^{1}}(\vec{p}_{d},\vec{p}_{o}) = 
    \begin{cases}
    	\eta_{1} \frac{\vec{p}_{do}}{\|\vec{p}_{do}\|^4}, & \|\vec{p}_{do}\| \leq P^{*} \\
    	0, & \|\vec{p}_{do}\| > P^{*}.
    \end{cases}
\end{equation}

A complete traditional potential field controller is the sum of (\ref{E.velocity_att}) and (\ref{E.velocity_rep_limit}) which yields (\ref{E.velocity_traditional}), where $n$ is the number of obstacles present in the environment.

\begin{equation}\label{E.velocity_traditional}
\resizebox{.9\columnwidth}{!}{$\displaystyle{
    \vec{v}_{d}^{PFC}(\vec{p}_{d},\vec{p}_{t},\vec{p}_{o}) =
    \begin{cases}
      -\lambda_{1}(\vec{p}_{t} - \vec{p}_{d}) + \sum_{i=0}^{n}\eta_{1} \frac{\vec{p}_{do}^{i}}{\|\vec{p}^i_{do}\|^4}, & \|\vec{p}^i_{do}\| \leq P^{*} \\
      -\lambda_{1}(\vec{p}_{t} - \vec{p}_{d}), & \|\vec{p}^i_{do}\| > P^{*}.
    \end{cases}
}$}
\end{equation}
\noindent

This controller enables a ground robot to track stationary or dynamic targets, while avoiding any obstacles in its path. However, when applied to an agile, aerial system such as a quadcopter, the controller's performance is quite poor as shown later in simulations presented in Section~\ref{S.simulation}.

\subsection{Extended Potential Field Controller}\label{S.extended_controller}

Because the potential field methods presented above are developed for ground robots, they do not address many of the factors that must be accounted for when designing a controller for aerial systems. For example, drones move very quickly and are inherently unstable which means they cannot simply move to a particular location and stop moving. They are consistently making fine adjustments to their position and velocity.

In order to account for factors unique to aerial platforms, this paper presents an extended potential field controller (ePFC) which utilizes the same concepts found in a traditional PFC, but applied to relative velocities rather than positions. Now, considering that the system is tracking a dynamic target, the desired velocity is that of the target. In this case, the attractive potential is defined as the quadratic function given by

\begin{equation}\label{E.attractive2}
    U_{att^{2}}(\vec{v}_{d},\vec{v}_{t}) = \frac{1}{2}\lambda_{2}\|\vec{v}_{dt}\|^{2},
\end{equation}

where $\lambda_{2}$ is positive scale factor, and $\|v_{dt}\|$ is the magnitude of the relative velocity between the drone velocity, $v_{d}$, and the target velocity, $v_{t}$, which is given by

\begin{equation}\label{E.rel_velocity}
    \|\vec{v}_{dt}\| = \sqrt{(\dot{x}_{dt})^{2}+(\dot{y}_{dt})^{2}+(\dot{z}_{dt})^{2}}.
\end{equation}

As in the traditional potential field controller the relative velocity potential should be minimized, thus resulting in a matched velocity between the drone and the target. Similar to the traditional controller, the desired velocity of the drone is found by calculating the negative gradient, which is

\begin{equation}\label{E.velocity_att2}
\begin{aligned}
    \vec{v}_{d}^{att^{2}}(\vec{v}_{d},\vec{v}_{t}) &= -\nabla U_{att^{2}}(\vec{v}_{d},\vec{v}_{t}),\\
    &= -\frac{\partial U_{att^{2}}}{\partial \dot{x}}\hat{i} - \frac{\partial U_{att^{2}}}{\partial \dot{y}}\hat{j} - \frac{\partial U_{att^{2}}}{\partial \dot{z}}\hat{k},\\
    &= -\nabla \Big(\frac{1}{2}\lambda_{2} \|\vec{v}_{dt}\|^{2})\Big),\\
    &= -\frac{1}{2}\lambda_{2}\nabla \|\vec{v}_{dt}\|^{2},\\
    &= -\frac{1}{2}\lambda_{2}\nabla (\dot{\vec{x}}_{dt}^{2}+\dot{\vec{y}}_{dt}^{2}+\dot{\vec{z}}_{dt}^{2}),\\
    &= -\lambda_{2} (\dot{\vec{x}}_{dt}+\dot{\vec{y}}_{dt}+\dot{\vec{z}}_{dt}),\\
    &= -\lambda_{2} (\vec{v}_{dt}).
\end{aligned}
\end{equation}

It is desirable that the drone and an obstacle should not maintain the same velocity, therefore the repulsive velocity potential between the drone and an obstacle is designed to be an inverse quadratic as in (\ref{E.repulsive}), given by

\begin{equation}\label{E.repulsive2}
    U_{rep^{2}}(\vec{v}_{d},\vec{v}_{o}) = \frac{1}{2}\eta_{2}\frac{1}{\|\vec{v}_{do}\|^{2}},
\end{equation}

\noindent
where $\eta_{2}$ is positive scale factor, and $\|\vec{v}_{do}\|$ is the magnitude of the relative velocity between the drone velocity, $\vec{v}_{d}$, and the obstacle velocity, $\vec{v}_{o}$. The corresponding velocity for this potential function is found by

\begin{equation}\label{E.velocity_rep2}
\begin{aligned}
    \vec{v}_{d}^{rep^{2}}(\vec{v}_{d},\vec{v}_{o}) &= \nabla U_{rep^{2}}(\vec{v}_{d},\vec{v}_{o}),\\
    &= \frac{\partial U_{rep^{2}}}{\partial \dot{x}}\hat{i} + \frac{\partial U_{rep^{2}}}{\partial \dot{y}}\hat{j} + \frac{\partial U_{rep^{2}}}{\partial \dot{z}}\hat{k},\\
    &= -\nabla \Big(\frac{1}{2}\eta_{2}\frac{1}{\|\vec{v}_{do}\|^{2}}\Big),\\
    &= -\frac{1}{2}\eta_{2} \nabla \frac{1}{\|\vec{v}_{do}\|^{2}},\\
    &= -\frac{1}{2}\eta_{2} \nabla \frac{1}{\dot{\vec{x}}_{do}^{2}+\dot{\vec{y}}_{do}^{2}+\dot{\vec{z}}_{do}^{2}},\\
    &= \eta_{2} \frac{\vec{v}_{d}-\vec{v}_{o}}{(\dot{\vec{x}}_{do}^{2}+\dot{\vec{y}}_{do}^{2}+\dot{\vec{z}}_{do}^{2})^{2}},\\
    &= \eta_{2} \frac{\vec{v}_{do}}{\|\vec{v}_{do}\|^{4}}.\\
\end{aligned}
\end{equation}

It should be noted that if the obstacle is stationary, then its velocity is zero. In this special case, the repulsive field in (\ref{E.repulsive2}) is designed to have no effect on the drone's motion. Therefore, the velocity found in (\ref{E.velocity_rep2}) becomes

\begin{equation}\label{E.velocity_rep2_limit}
    \vec{v}_{d}^{rep^{2}}(\vec{v}_{d},\vec{v}_{o}) = 
    \begin{cases}
    	\eta_{2} \frac{\vec{v}_{d}-\vec{v}_{o}}{\|\vec{v}_{do}\|^{4}}, & \|\vec{v}_{o}\| \neq 0 \\
    	0, & \|\vec{v}_{o}\| = 0.
    \end{cases}
\end{equation}

\begin{figure}[htb]
 \centering
  \includegraphics[width=1\columnwidth]{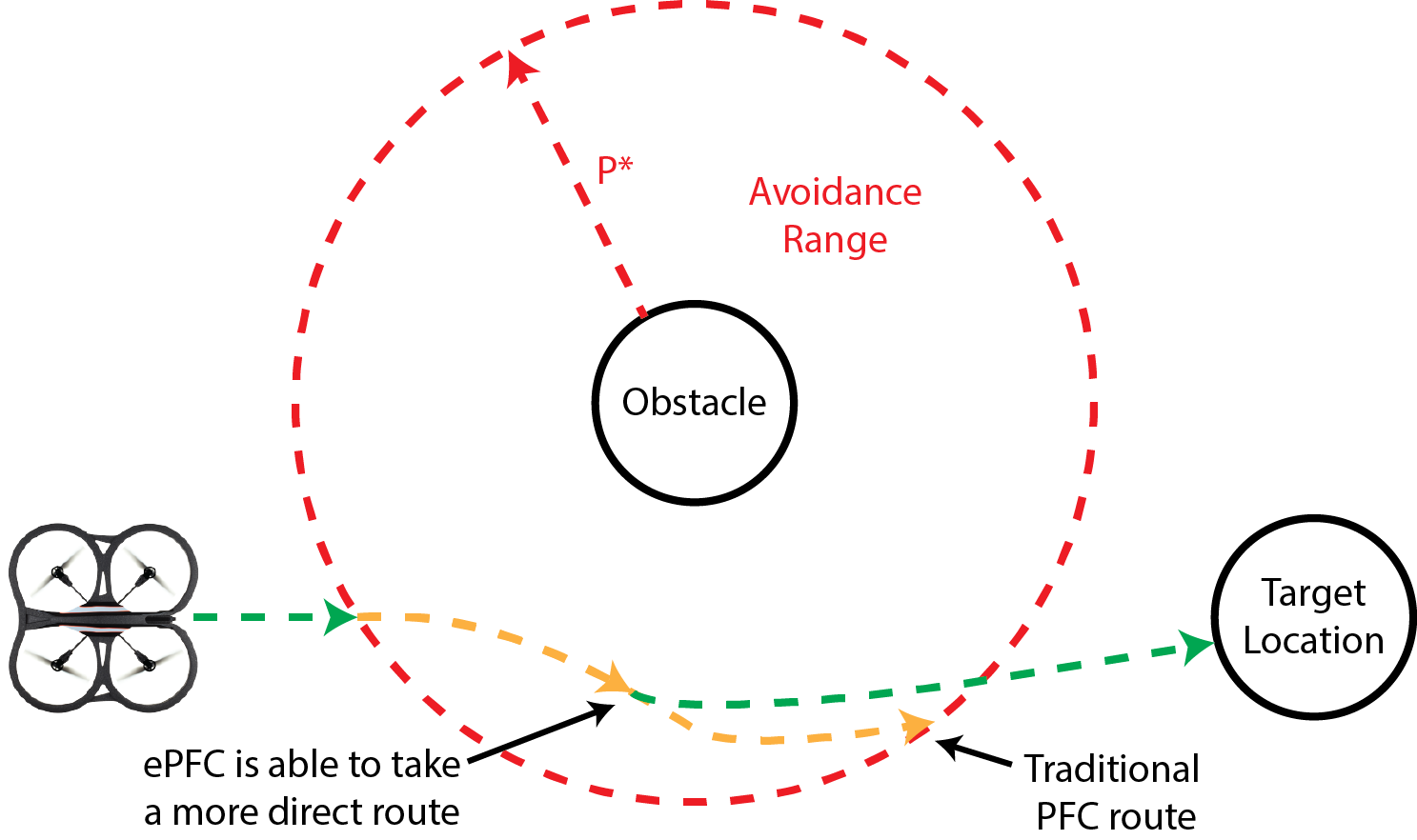}
  \caption{By taking into account the time rate of change of $\|\vec{p}_{do}\|$, the controller is able to ignore repulsive effects from the obstacle if the drone is moving away from the obstacle ($\dot{\|p_{do}\|} \geq 0$). This results in a more direct route to the target, thus saving time.}
  \label{F.repulsive_limit2}
\end{figure}

Finally, the relative distance between the drone and obstacle, $\vec{p}_{do}$, is revisited. In the traditional potential field controller presented previously, $\vec{p}_{do}$ was used as the basis for a repulsive potential. However, no thought is given to the time rate of change of the magnitude of $\vec{p}_{do}$. If a situation in which the drone is moving away from the obstacle is considered, then $\dot{\|p_{do} \|} \geq 0$ in which case no avoidance action needs to be taken, even if the drone is within the avoidance range. As illustrated in Fig.~\ref{F.repulsive_limit2}, the controller is able to ignore the effect of the obstacle sooner, and therefore can take a more direct route, thus saving time.

Furthermore, the drone should take evasive action if $\dot{\|p_{do}\|} < 0$, when the distance between the two is decreasing. Therefore, a final control effort is designed as

\begin{equation}\label{E.velocity_rep3_limit}
    \vec{v}_{d}^{rep^{3}}(\vec{p}_{d},\vec{p}_{o}) = 
    \begin{cases}
    	-\eta_{3} \dot{\|p_{do}\|} \frac{\vec{p}_{do}}{\|\vec{p}_{do}\|}, & \dot{\|p_{do}\|} < 0 \\
    	0, & \dot{\|p_{do}\|} \geq 0,
    \end{cases}
\end{equation}

\noindent
where $\eta_{3}$ is positive scale factor.

Summing the velocities in (\ref{E.velocity_att2}), (\ref{E.velocity_rep2_limit}), and (\ref{E.velocity_rep3_limit}) with the traditional controller (\ref{E.velocity_traditional}) yields the full form of the extended potential field controller (ePFC), which is

\begin{equation}\label{E.velocity_complete}
    \vec{v}_{d}^{ePFC} = \vec{v}_{d}^{PFC}-\lambda_{2}(\vec{v}_{d} - \vec{v}_{t}) + \sum_{i=0}^{n}\eta_{2} \frac{\vec{v}_{d}-\vec{v}_{o}^{i}}{\|\vec{v}_{do}^{i}\|^{4}} - \sum_{i=0}^{n}\eta_{3} \dot{\|p_{do}^{i}\|} \frac{\vec{p}_{do}^{i}}{\|\vec{p}_{do}^{i}\|},
\end{equation}

\noindent
where $n$ is the number of obstacles present, $\|v_{o}^{i}\| \neq 0$, $\dot{\|p_{do}^{i}\|} < 0$, and the same conditions discussed previously apply to $v_d^{PFC}$.

Finally, the velocity found in (\ref{E.velocity_complete}) must be transformed into the body coordinate system of the drone and is found to be

\begin{equation}\label{E.body_velocity}
\begin{aligned}
   \vec{v}_{d,body}^{ePFC} &= \vec{v}_{d}^{ePFC}*
   \begin{bmatrix}
    cos(\psi)       & -sin(\psi) & 0 \\
    sin(\psi)       &  sin(\psi) & 0 \\
    0           &  0     & 1 
  \end{bmatrix},
\end{aligned}
\end{equation}

\noindent
where $\psi$ is the yaw angle of the drone around the body $z$ axis.

This controller seeks out a moving target, and also avoids obstacles that are in close proximity.

\begin{figure*}[htb]
 \centering
  \includegraphics[width=0.85\textwidth]{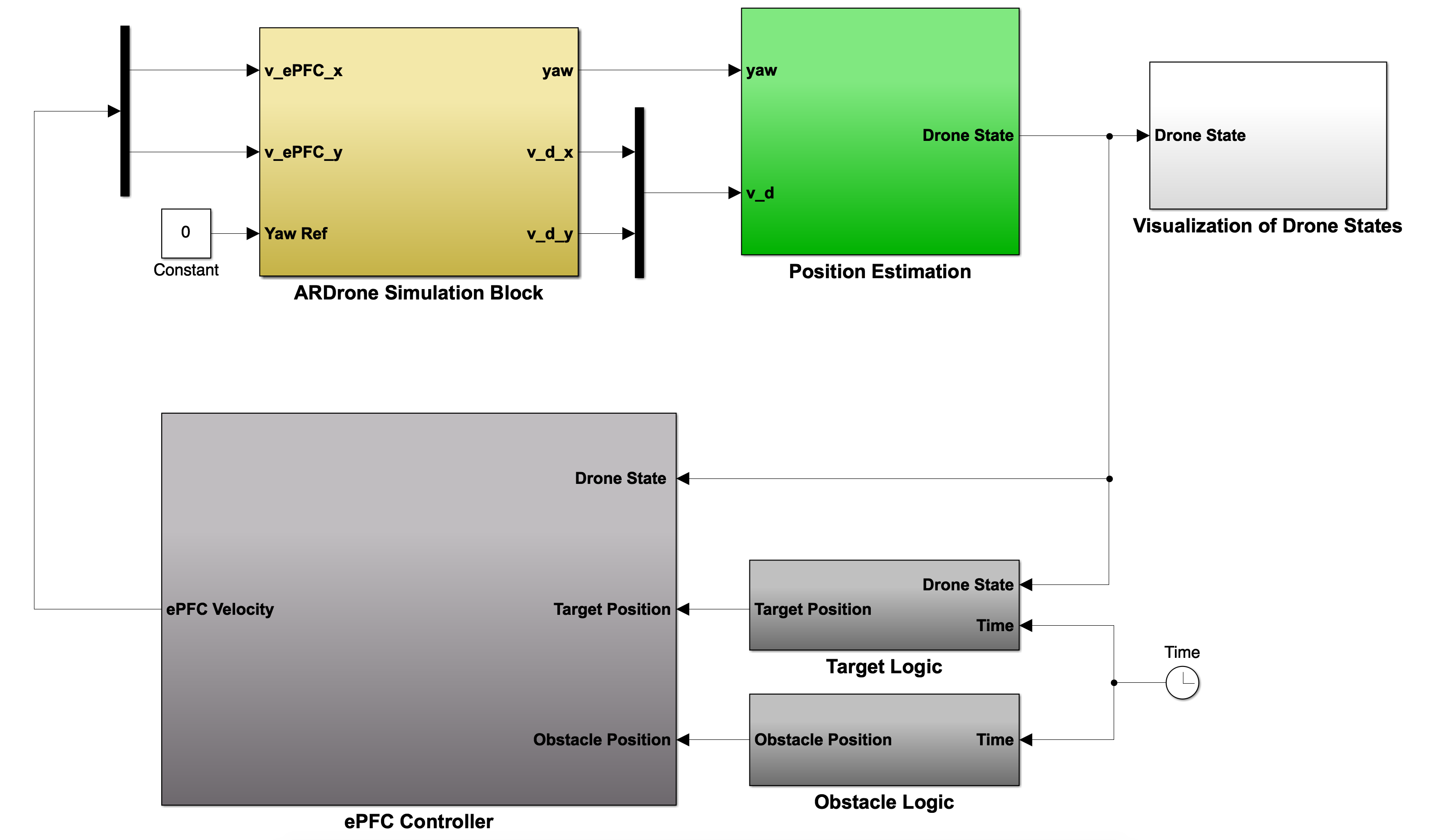}
  \caption{The Simulink model used includes a state space representation from the ARDrone Simulink Development Kit, as well as custom blocks for the ePFC controller described in this work.}
  \label{F.simulink_overview}
\end{figure*}

\subsection{Stability Analysis}\label{S.lyapunov}

To analyze the convergence of the proposed velocity controller (\ref{E.velocity_complete}) for the drone, the Lyapunov theory is used. Consider a positive definite Lyapunov function as follows:

\begin{equation}\label{E.Lyapunov1}
  L = U_{att} = \frac{1}{2}\lambda_{1} \|\vec{p}_{dt}\|^{2} + \frac{1}{2}\lambda_{2} \|\vec{v}_{dt}\|^{2}.
\end{equation}

This function represents the artificial potentials of the controller. Since (\ref{E.Lyapunov1}) is positive definite, its Lie derivative is given by

\begin{equation}\label{E.Lyapunov_dot1}
\begin{aligned}
  L^{*} &= \frac{\partial L}{\partial \vec{p}_{dt}} \vec{v}_{dt} + \frac{\partial L}{\partial \vec{v}_{dt}} \vec{a}_{dt},\\
  &= \lambda_{1} \|\vec{p}_{dt}\| \vec{v}_{dt} + \lambda_{2} \|\vec{v}_{dt}\| \vec{a}_{dt},\\
\end{aligned}
\end{equation}

\noindent
where $\vec{a}_{dt}$ is the relative acceleration between the drone acceleration and the target acceleration.

Note that the relative velocity between the drone and the target is designed following the direction of the negative gradient of $U_{att}(p_{dt})$ with respect to $p_{dt}$ as in (\ref{E.velocity_att}). From (\ref{E.velocity_att2}), obtain

\begin{equation}\label{E.acceleration}
\begin{aligned}
  \vec{a}_{dt} &= \dot{\vec{v}}_{dt} = \frac{d}{dt}\left(-\frac{1}{2}\lambda_{2}\nabla(\dot{\vec{x}}_{dt}^{2}+\dot{\vec{y}}_{dt}^{2}+\dot{\vec{z}}_{dt}^2)\right),\\
  &= \frac{d}{dt}\left(-\lambda_{2}\|\vec{v}_{dt}\|\right),\\
  &= - \lambda_{2} \| \frac{\vec{v}_{dt}(t) - \vec{v}_{dt}(t-1)}{\Delta_t} \|,\\
\end{aligned}
\end{equation}

\noindent
where $\Delta_t$ is a time step. Hence, substituting $v_{dt}$ given by (\ref{E.velocity_att}) and $a_{dt}$ given by (\ref{E.acceleration}) into (\ref{E.Lyapunov_dot1}), 

\begin{equation}\label{Lyapunov_dot2}
  L^{*} = -\left[\lambda_{1}^{2} \|\vec{p}_{dt}\|^{2} + \lambda^2_2 \|\vec{v}_{dt}\| \|\frac{\vec{v}_{dt}(t) - \vec{v}_{dt}(t-1)}{\Delta_t}\|\right].
\end{equation}

It can be easily seen that $L^{*} < 0$ since $\|\vec{p}_{dt}\|$, $\|\vec{v}_{dt}\|$, and $ \|\frac{\vec{v}_{dt}(t) - \vec{v}_{dt}(t-1)}{\Delta_t}\|$ are positive. This means that the proposed controller is stable, and the drone is able to track a moving target.

\section{Simulation}
\label{S.simulation}
\subsection{MATLAB Environment}
In order to validate the developed controller, the system was simulated using a Matlab Simulink model. The state space representation of the ARDrone's platform dynamics are take from the ARDrone Simulink Development Kit~\cite{Sanabaria_link}. The complete Simulink model shown in Fig.~\ref{F.simulink_overview} demonstrates how the ePFC controller uses feedback information from the ARDrone simulation and position estimator blocks. The output of the ARDrone simulation block is simply the velocity of the drone, and the position estimator uses an integrator with zero initial conditions to calculate position.

The desired path that the drone is to take is outlined in Table~\ref{T.waypoints}. A virtual obstacle is placed at $(1,1)$ which places it immediately in the path of the drone between waypoints 2 and 3. The drone is allowed two seconds at each waypoint in an attempt to let it settle before moving on to the next waypoint.

\begin{table}[htb]
\renewcommand{\arraystretch}{1.2}
\caption{Simulation Waypoints}
\label{T.waypoints}
\centering
\begin{tabularx}{1\columnwidth}{@{\extracolsep{\stretch{1}}}*{3}{c}@{}}

\bfseries Waypoint & \bfseries X Coordinate [m] & \bfseries Y Coordinate [m]\\
\hline
\hline
1 &  2.5  & -1 \\
2 &  2.5  &  1 \\
3 & -2.5  &  1 \\
4 & -2.5  & -1 \\
\hline
\end{tabularx}
\end{table}

\subsection{Simulation results}
\begin{figure}[H]
 \centering
  \includegraphics[width=\columnwidth]{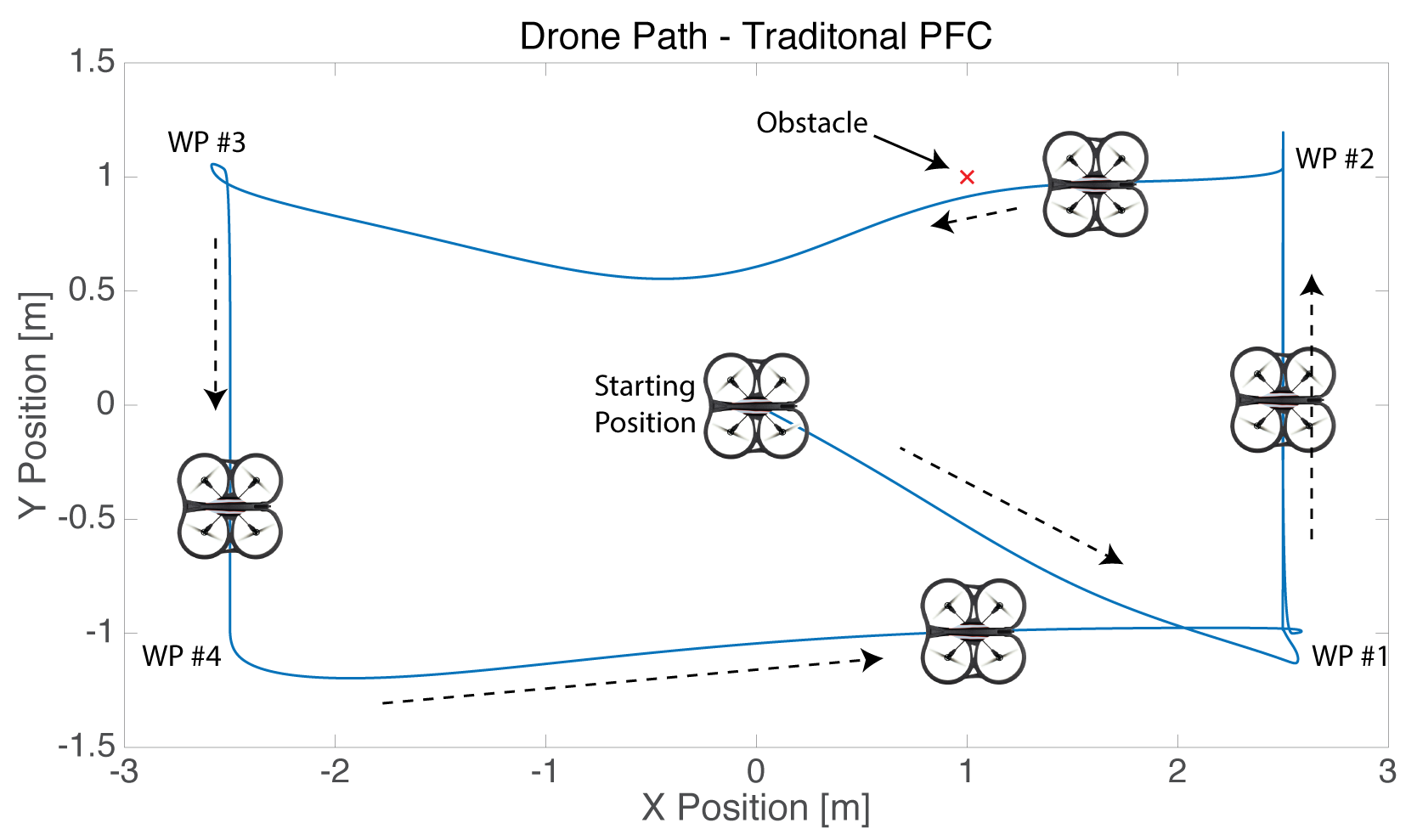}
  \caption{A traditional PFC is simulated on the ARDrone, with poor results. Because drones cannot stop instantaneously like ground robots, the drone often overshoots the desired waypoint. For reference, the drone takes approximately 35 seconds to complete a full loop of the course.}

  \label{F.traditional_sim_path}
    \vspace{-10pt}
\end{figure}

First, a traditional potential field controller was simulated, and the resulting path is shown in Fig.~\ref{F.traditional_sim_path}. The performance of the traditional PFC was poor as expected, because aerial drones have very different dynamics than their ground counterparts. Using the traditional PFC, the drone overshoots the desired waypoint, and while it does avoid the obstacle at $(1,1)$ it is not by much. The drone completed a full loop in approximately 35 seconds.

\begin{figure}[H]
 \centering
  \includegraphics[width=\columnwidth]{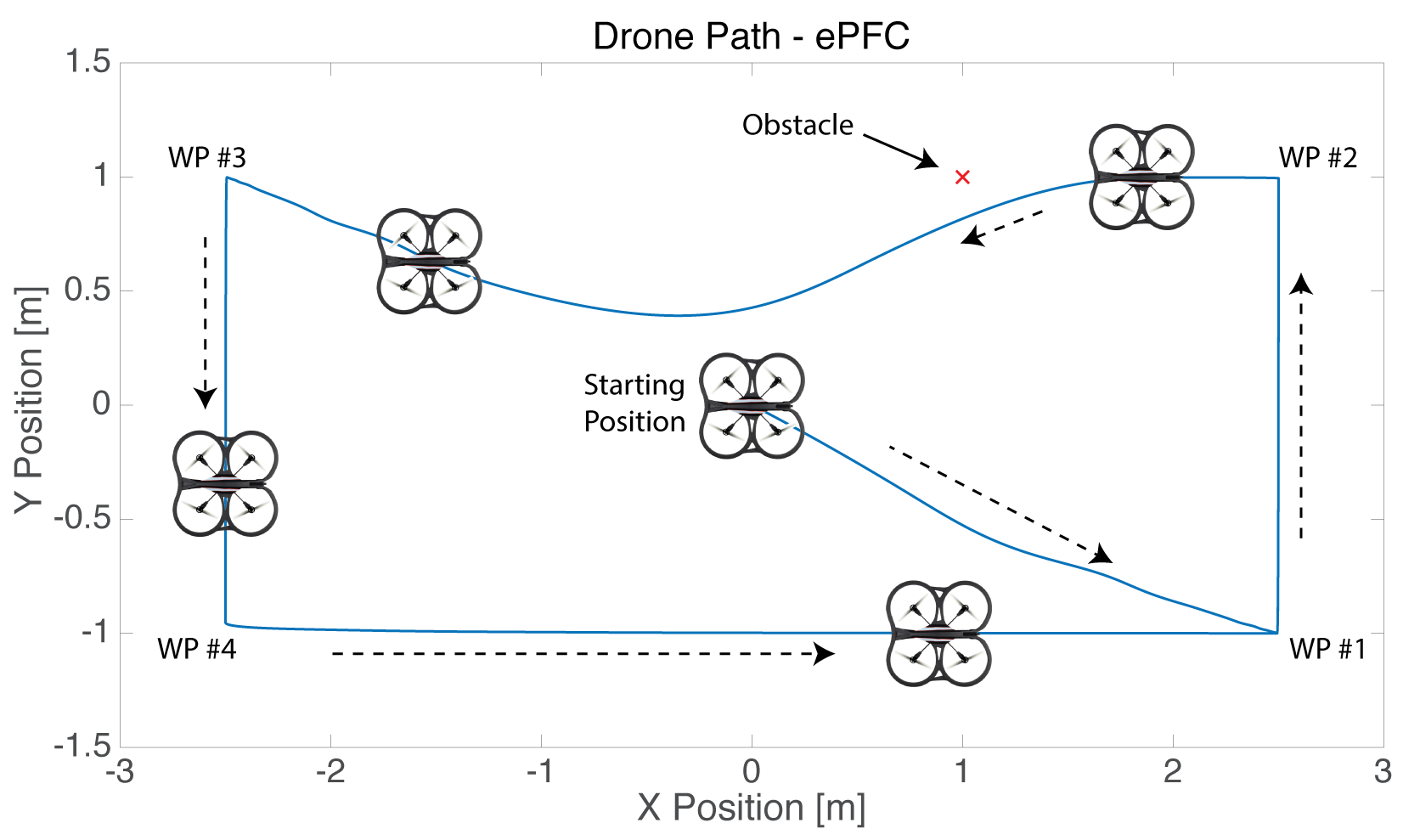}
  \caption{Using the extended potential field controller (ePFC), the drone is able to complete the course without overshooting the target waypoints, and avoids the obstacle by a larger margin than the traditional controller. Because the drone does not overshoot the target, it is able to complete the course in a shorter amount of time compared to the traditional controller.}
  \label{F.extended_sim_path}
  \vspace{-0pt}
\end{figure}

Next, the ePFC is tested using the same path and obstacle position. The results shown in Fig.~\ref{F.extended_sim_path} demonstrate the effectiveness of the new controller. The drone does not overshoot the desired waypoints and avoids the obstacle by a larger margin, while completing the course in a shorter amount of time than the traditional controller.

\begin{table}[htb]
\renewcommand{\arraystretch}{1.2}
\caption{Simulation Controller Evaluation}
\label{T.sim_results}
\centering
\begin{tabularx}{1\columnwidth}{@{\extracolsep{\stretch{1}}}*{3}{c}@{}}

\bfseries Controller & \bfseries Overshoot [\%]& \bfseries Settling Time [sec]\\
\hline
\hline
Traditional PFC &  \textgreater19\%  & 6 \\
ePFC &  0\%  &  5 \\
\hline
\end{tabularx}
\end{table}

As outlined in Table~\ref{T.sim_results}, the ePFC controller has zero overshoot, and has a settling time of approximately five seconds. This is a large improvement over the traditional controller which overshoots by up to $19\%$ and takes nearly six seconds to settle. It is clear that the proposed controller is more appropriate for use on an aerial drone than the traditional PFC. It is also important to note that at the furthest point (waypoint 4) the drone is approximately $4$m from the obstacle which is still within the avoidance range P* for the simulation. The ePFC outperforms the traditional PFC because it takes into account the changing relative distance between the drone and the obstacle behind it. Since the relative distance between the drone and the obstacle is increasing, the repulsive obstacle potential has no effect as mentioned in (35). The traditional PFC does not take this into account and therefore is being pushed past the target even though it is already moving away from it.

\begin{figure}[H]
 \centering
  \includegraphics[width=\columnwidth]{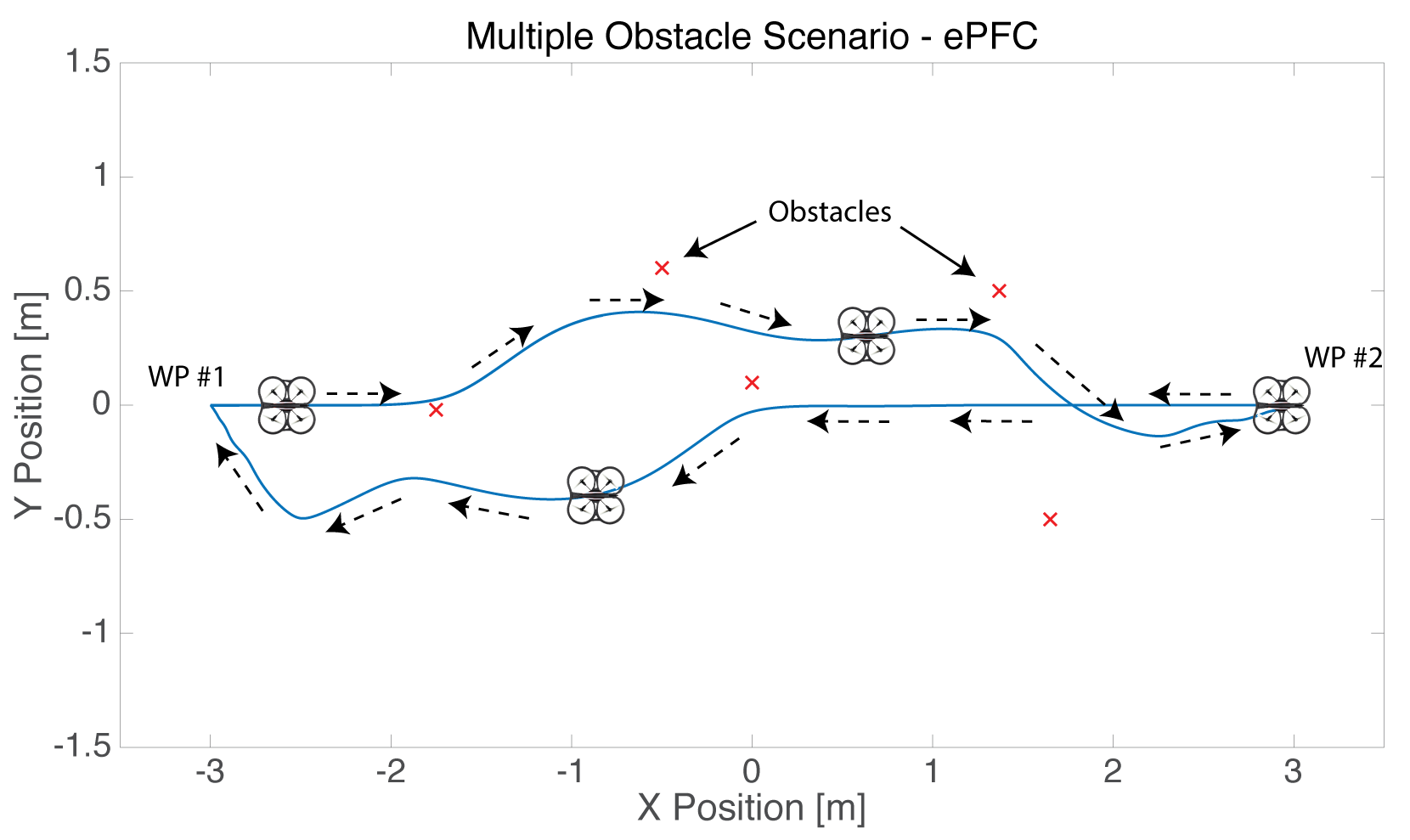}
  \caption{A more complex simulation was performed with multiple obstacles placed throughout the environment. The drone is able to successfully navigate between waypoints without colliding with a single obstacle, thus demonstrating its effectiveness in multi-obstacle scenarios.}
  \label{F.complex_sim_path}
\end{figure}

In addition to the comparison between the tradition PFC and the ePFC, a more complex simulation was performed which included several obstacles placed at random throughout the environment. The results shown in Fig.~\ref{F.complex_sim_path} demonstrate the that the ePFC is very effective in multi-obstacle scenarios, and it successfully navigates between waypoints without colliding with a single obstacle.

\section{Experimental Results}\label{S.Experiment}

This section presents the experimental setup used to implement the proposed controller. In addition, the results from implementation are presented and the performance of the proposed controller is discussed.

\begin{figure}[t]
 \centering
  \includegraphics[width=\columnwidth]{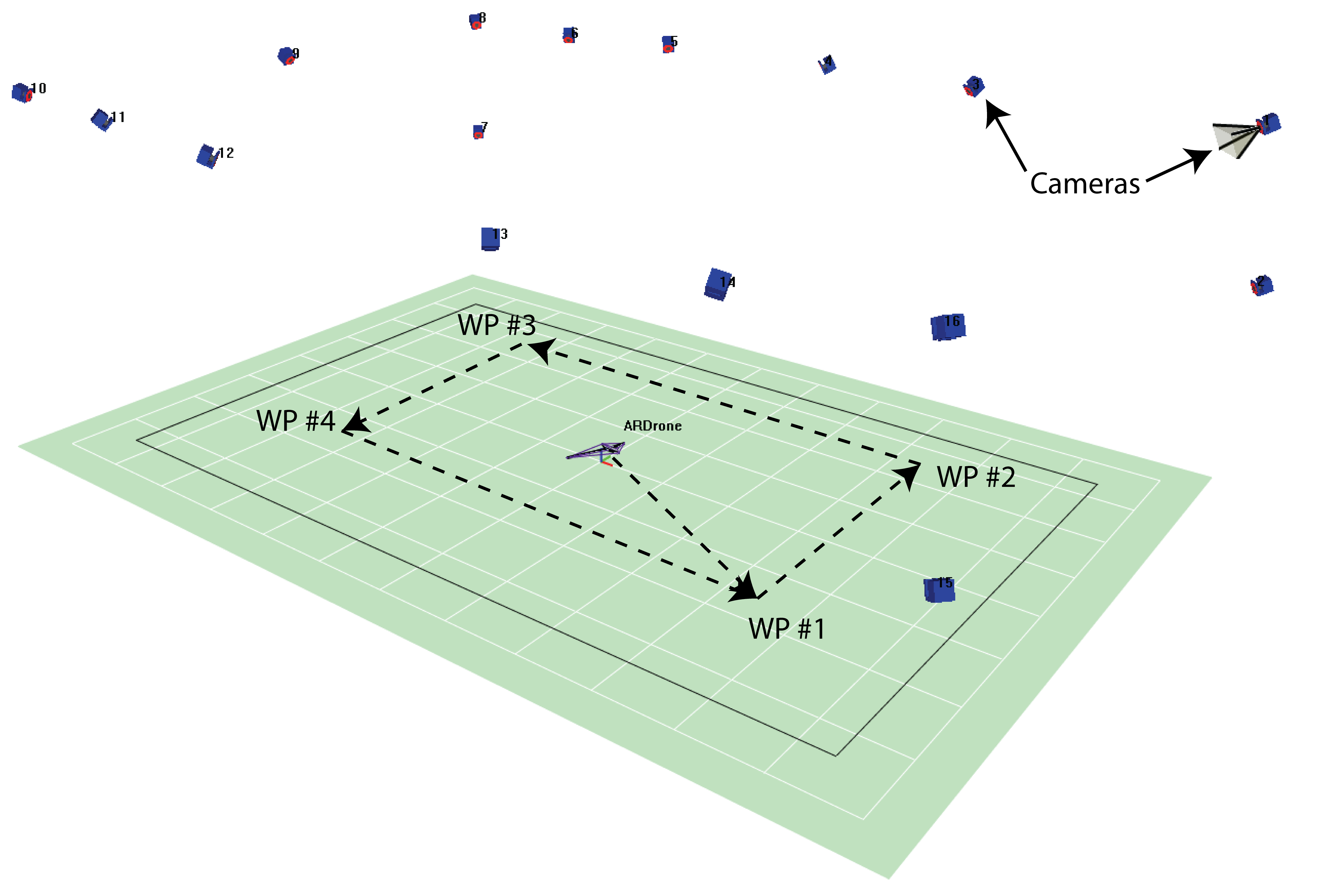}
  \caption{The Motion Analysis Cortex software gives the user a real-time, visual 3D representation of the environment including camera locations and any objects sensed by the system~\cite{MotionCap}.}
  \label{F.cortex_overview}
\end{figure}

\begin{figure*}[t]
 \centering
  \includegraphics[width=0.85\textwidth]{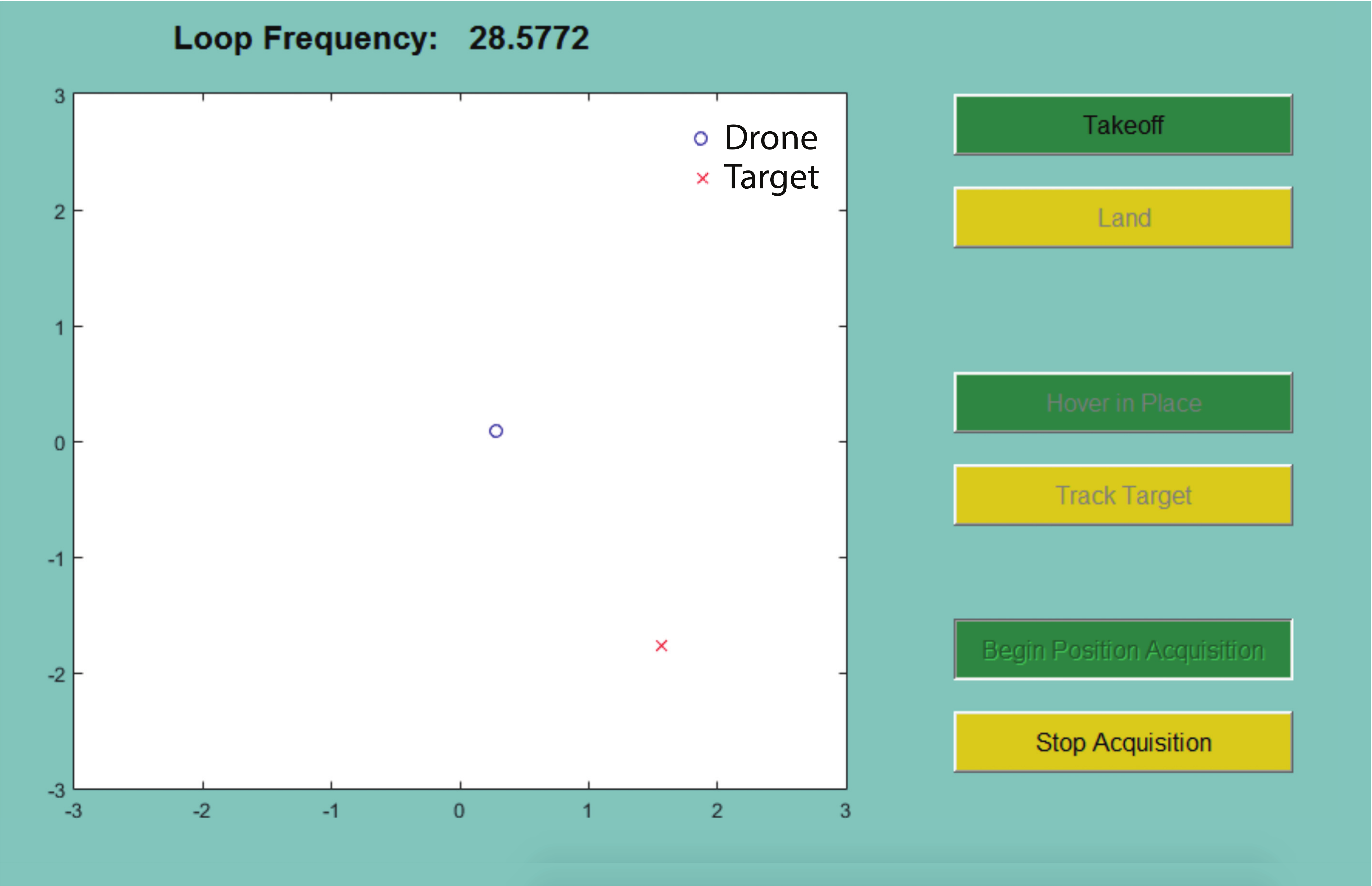}
  \caption{A Matlab GUI was created to show the positions of the drone, target, and any obstacles present. It also allows the user to control when the drone takes off, lands, tracks the target, or simply hovers in place.}
  \label{F.gui}
\end{figure*}

\subsection{Experimental Setup}

The experimental platform chosen to implement the ePFC is the ARDrone 2.0 quadcopter shown in Fig.~\ref{F.ARDrone}. This platform was chosen for its ease of communication - over a wifi connection - as well as the safety provided by the foam hull. Additionally, the ARDrone requires little to no setup and spare parts are readily available in case of crashes. The ARDrone 2.0 can be equipped with a $1500$ mAh battery which yields flight times up to $18$ min. Large batteries and long flight times are very advantageous in a testing environment because it allows for more uninterrupted tests and less downtime recharging batteries. The ARDrone 2.0 is also equipped with a $1$ GHz 32 bit ARM Cortex A8 processor, $1$ GB DDR2 RAM, and runs Linux. This means that the developed controller can be implemented on-board the drone in future work.

Sixteen Motion Analysis Kestrel cameras located throughout the testing space provide the position and orientation of the drone, target (if not virtual), and any obstacles present. The Cortex software suite provides a visual representation of the environment as shown in Fig.~\ref{F.cortex_overview} as well as sending data over a network connection for use by external programs.

In order to control the drone and display its location along with the target and any obstacles present, the Matlab GUI shown in Fig.~\ref{F.gui} was created. It allows the user to determine when the drone takes off, lands, or tracks the target. This GUI is critical in efficient testing of the drone. Additionally, for the safety of the drone, if the controller does not behave as expected the user can request that the drone simply hover in place to avoid fly-aways.

\begin{figure}[htb]
 \centering
  \includegraphics[width=0.75\columnwidth]{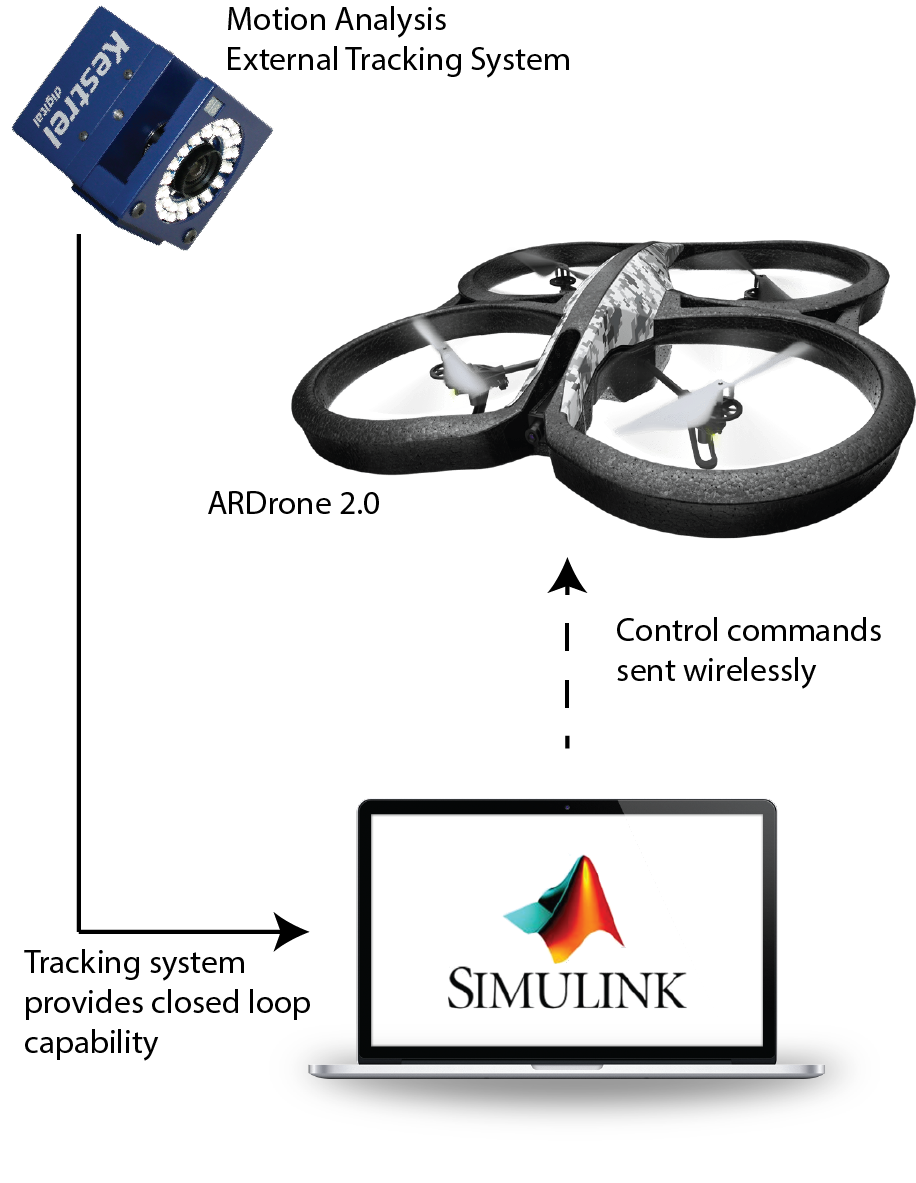}
  \caption{The experimental setup for this work includes a Motion Analysis external tracking system with 16 cameras which provides the position of the drone and obstacles in the environment. The same Simulink model used in Section~\ref{S.simulation} provides control commands to the ARDrone over a wireless connection.}
  \label{F.system}
\end{figure}

The overview of the experimental setup shown in Fig.~\ref{F.system} demonstrates the feedback loop implemented. The Motion Analysis external tracking system is used for localization of the drone and obstacles in real time. The position information is used by the same Simulink model shown in Section~\ref{S.simulation} which controls the ARDrone over a wireless connection. 

\subsection{Experimental Results and Discussion}

Having validated the controller using the Simulink simulation, it was then implemented on the actual ARDrone. Several experiments were formed, in the order outlined in Table~\ref{T.exp_tests}.

\begin{table}[htb]
\renewcommand{\arraystretch}{1.2}
\caption{Experimental Tests}
\label{T.exp_tests}
\centering
\begin{tabularx}{1\columnwidth}{@{\extracolsep{\stretch{1}}}*{3}{c}@{}}

\bfseries Test Number & \bfseries Target & \bfseries Obstacle(s)\\
\hline
\hline
1 & 1 - Static  & 0 - N/A \\
2 & 1 - Static  & 1 - Static \\
3 & 1 - Dynamic Square & 0 - N/A \\
4 & 4 - Static Waypoints  &  1 - Static \\
5 & 1 - Dynamic No Pattern & 0 - N/A \\
\hline
\end{tabularx}
\end{table}

Because the simulation showed a clear improvement in performance between the tradition PFC and the developed ePFC, the traditional controller was not tested on the experimental platform. Instead, the ePFC was immediately implemented in the experiments.

\begin{figure}[htb]
\centering
\includegraphics[width=1\columnwidth]{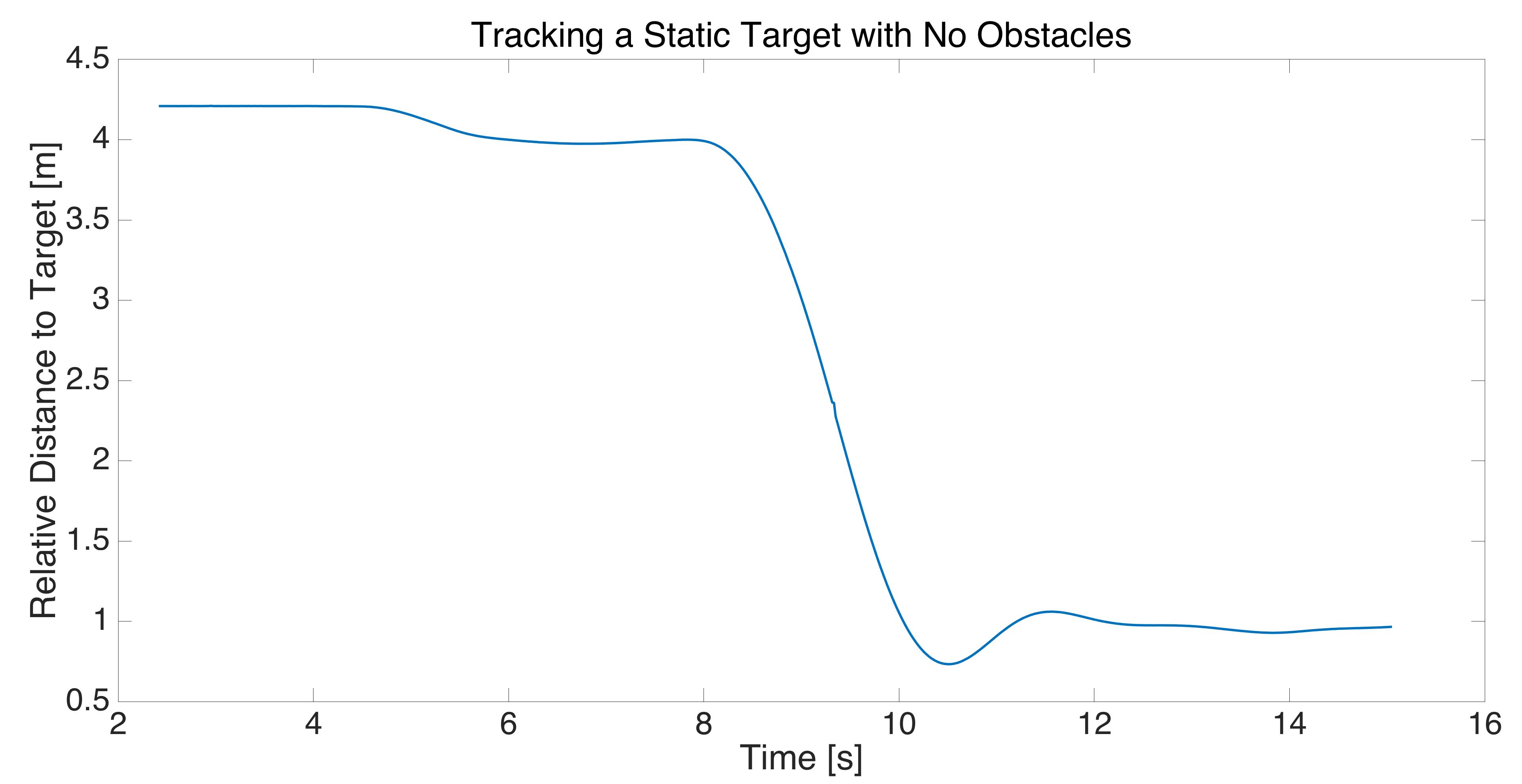}\\
\caption{The results of tracking a static target with no obstacles are very good. With an initial condition of approximately $3.2$m, the drone achieves position in under $5$sec.}
\label{F.staticNoObstacles}
\end{figure}

In the first test, the drone was placed approximately $4.2$m from the target's location. Because the target wand is often held by a human, the drone was requested to fly to $1$m away from the target location to avoid collision with someone holding the wand. The drone's response shown in Fig.~\ref{F.staticNoObstacles} demonstrates the capability of the drone to achieve a goal position effectively. Starting at approximately $7.75$sec, the drone enters an autonomous mode, and achieves stable hover $1$m away from the target in approximately $5$sec. It is important to note that while the drone did overshoot it's goal location, it did not overshoot enough to get close to hitting the target. The closest that the drone got to the target was just under $0.75$m.

\begin{figure}[htb]
\centering
\includegraphics[width=1\columnwidth]{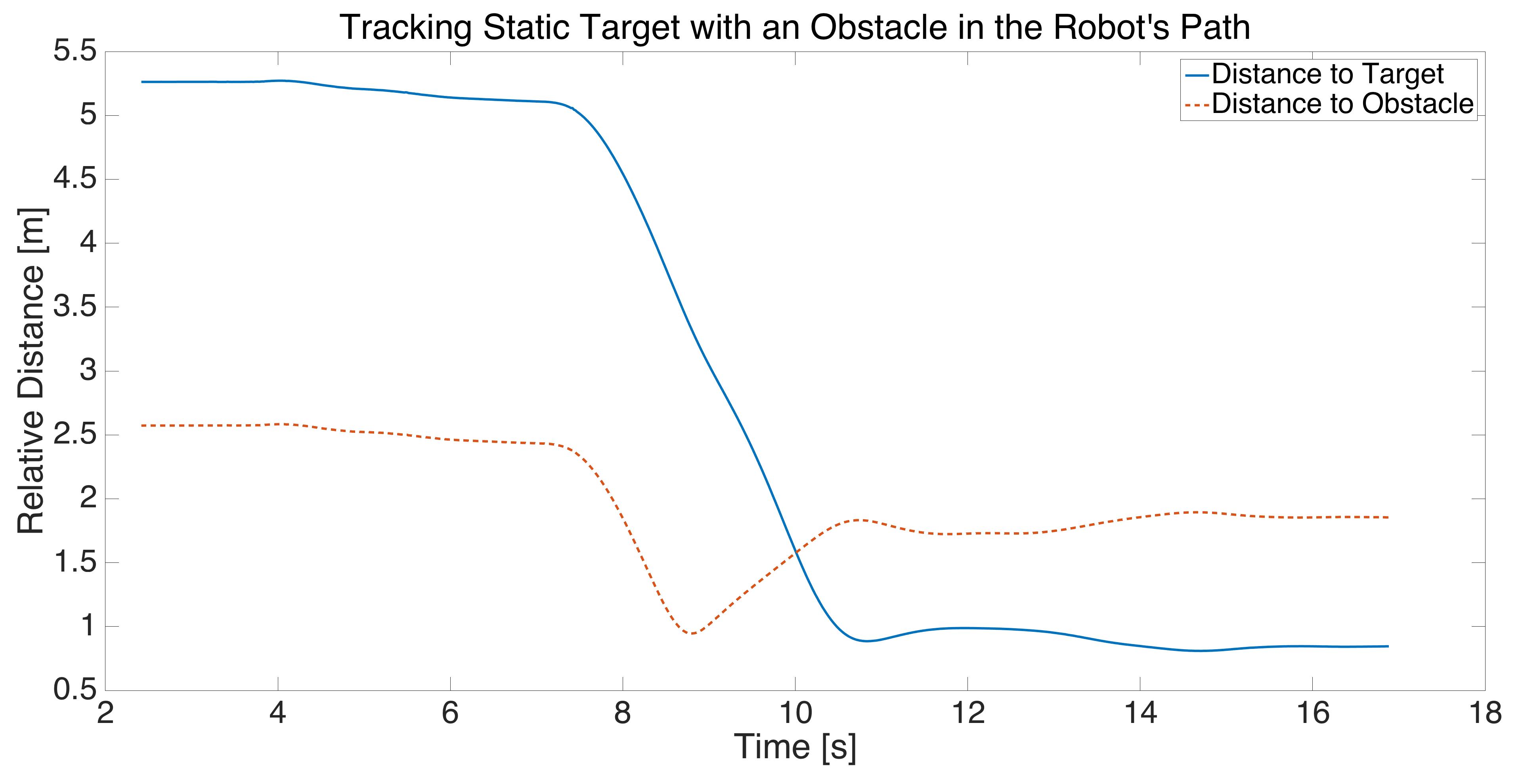}\\
\caption{The results of tracking a static target with an obstacle in the way demonstrates the controllers effectiveness at avoiding collisions. As expected, the drone's position relative to the obstacle decreases, but the drone takes avoidance action and never gets closer than one meter away from the obstacle.}
\label{F.staticWithObstacles1}
\vspace{-5pt}
\end{figure}
In the second experiment, the drone was placed approximately $5.2$m away from the target wand, and an obstacle was located in the path lying directly between the drone and the target. Similar to the first test, the drone's mission was to fly to within $1$m of the target, this time while avoiding the obstacle and still achieving the task. As the drone begins moving towards the target, it also moves towards the obstacle. Because of the repulsive forces generated by the relative position and velocity with respect to the obstacle, the drone is elegantly pushed around the obstacle and still makes it to the target location. Figure~\ref{F.staticWithObstacles1} shows the results of this test, demonstrating that the drone maintains a safe distance from the obstacle ($1$m minimum) and also achieves the goal.

\begin{figure}[htb]
\centering
\includegraphics[width=1\columnwidth]{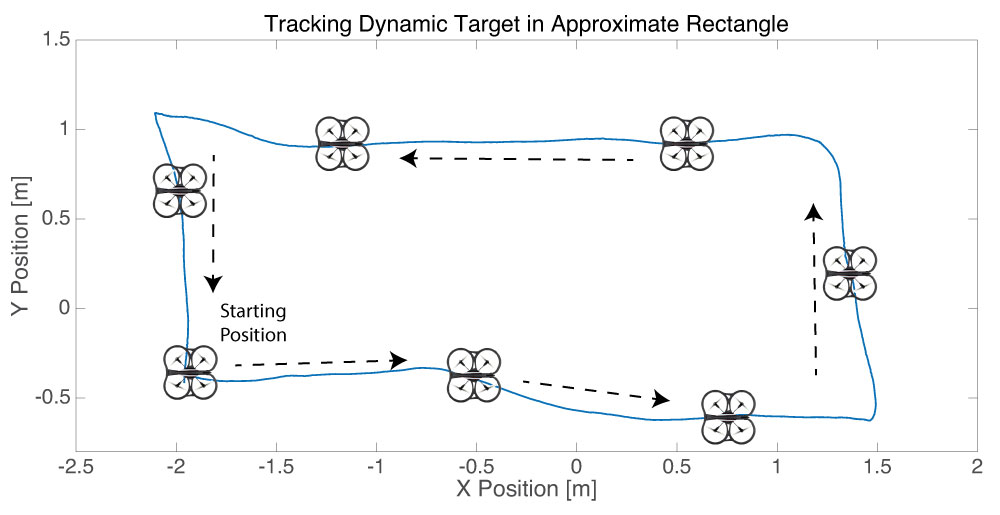}\\
\caption{In the third experiment, the drone tracks a target which moves in an approximate rectangle. As shown, the drone does track, but because the desired trajectory is human-controlled the reference is not perfect. Test number four addresses this imperfection by using set virtual waypoints.}
\label{F.square}
\end{figure}

In the third test, the drone was instructed to follow the target wand as it moved in an approximate rectangle around the lab. The results shown in Fig.~\ref{F.square} illustrate the path of the drone as it follows the target through the pattern. As shown, the drone does in fact track the rectangle as instructed. Because the path of the target was moved manually by a person holding the wand, the target trajectory is not a perfect rectangle. Therefore, the next experiment establishes a perfect rectangle using virtual waypoints.

In test number four, virtual waypoints like those used in simulation are used to demonstrate the ability of the drone to navigate a course and avoid obstacles. The waypoints used for this test are outlined in Table~\ref{T.exp_waypoints}.

\begin{table}[htb]
\renewcommand{\arraystretch}{1.2}
\caption{Experimental Waypoints}
\label{T.exp_waypoints}
\centering
\begin{tabularx}{1\columnwidth}{@{\extracolsep{\stretch{1}}}*{3}{c}@{}}

\bfseries Waypoint & \bfseries X Coordinate [m] & \bfseries Y Coordinate [m]\\
\hline
\hline
1 &  1.5  & -0.5 \\
2 &  1.5  &  0.5 \\
3 & -1.5  &  0.5 \\
4 & -1.5  & -0.5 \\
\hline
\end{tabularx}
\end{table}

\begin{figure}[htb]
 \centering
  \includegraphics[width=1\columnwidth]{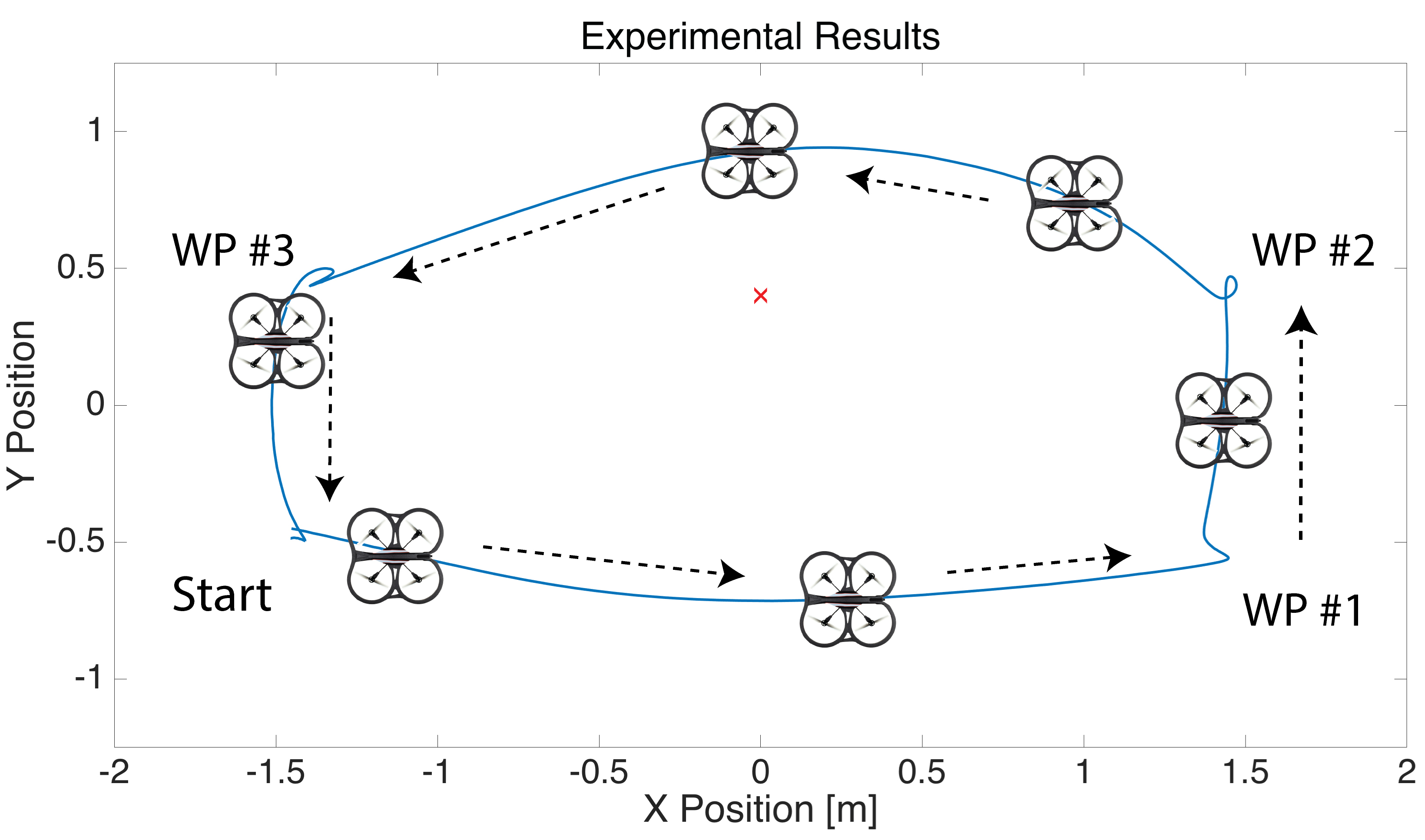}
  \caption{To test the ePFC experimentally, the drone follows waypoints similar to those in the simulation. The drone successfully reaches each waypoint and avoids the obstacle in its path between waypoints two and three. }
  \label{F.exp_results1}
\end{figure}

\begin{figure}[htb]
 \centering
  \includegraphics[width=1\columnwidth]{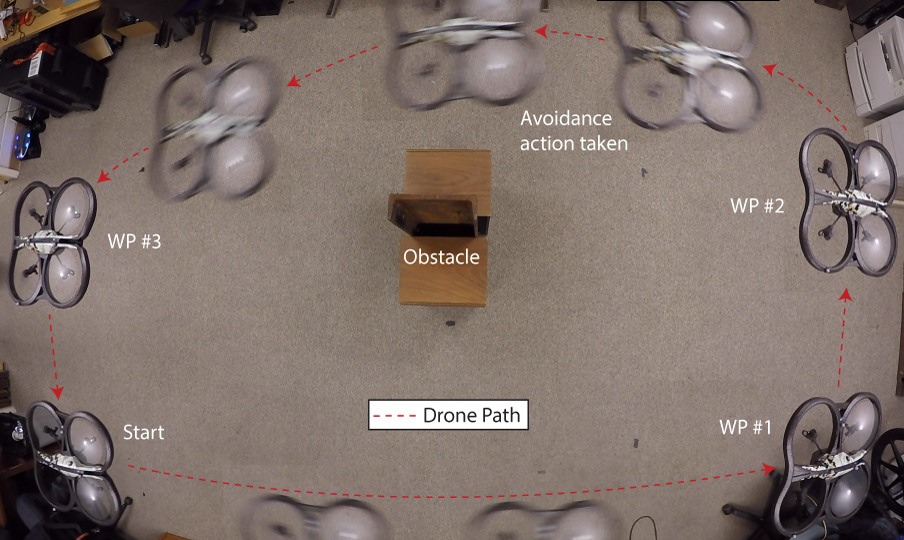}
  \caption{Stills from a video demonstrate the drone taking avoidance action while tracking moving waypoints.}
  \label{F.exp_results2}
\end{figure}

The results from the fourth experiment shown in Fig.~\ref{F.exp_results1} and Fig.~\ref{F.exp_results2} demonstrate that the drone successfully reaches each waypoint, and also avoids the obstacle in its path between waypoints two and three.

\begin{figure}[htb]
 \centering
  \includegraphics[width=1\columnwidth]{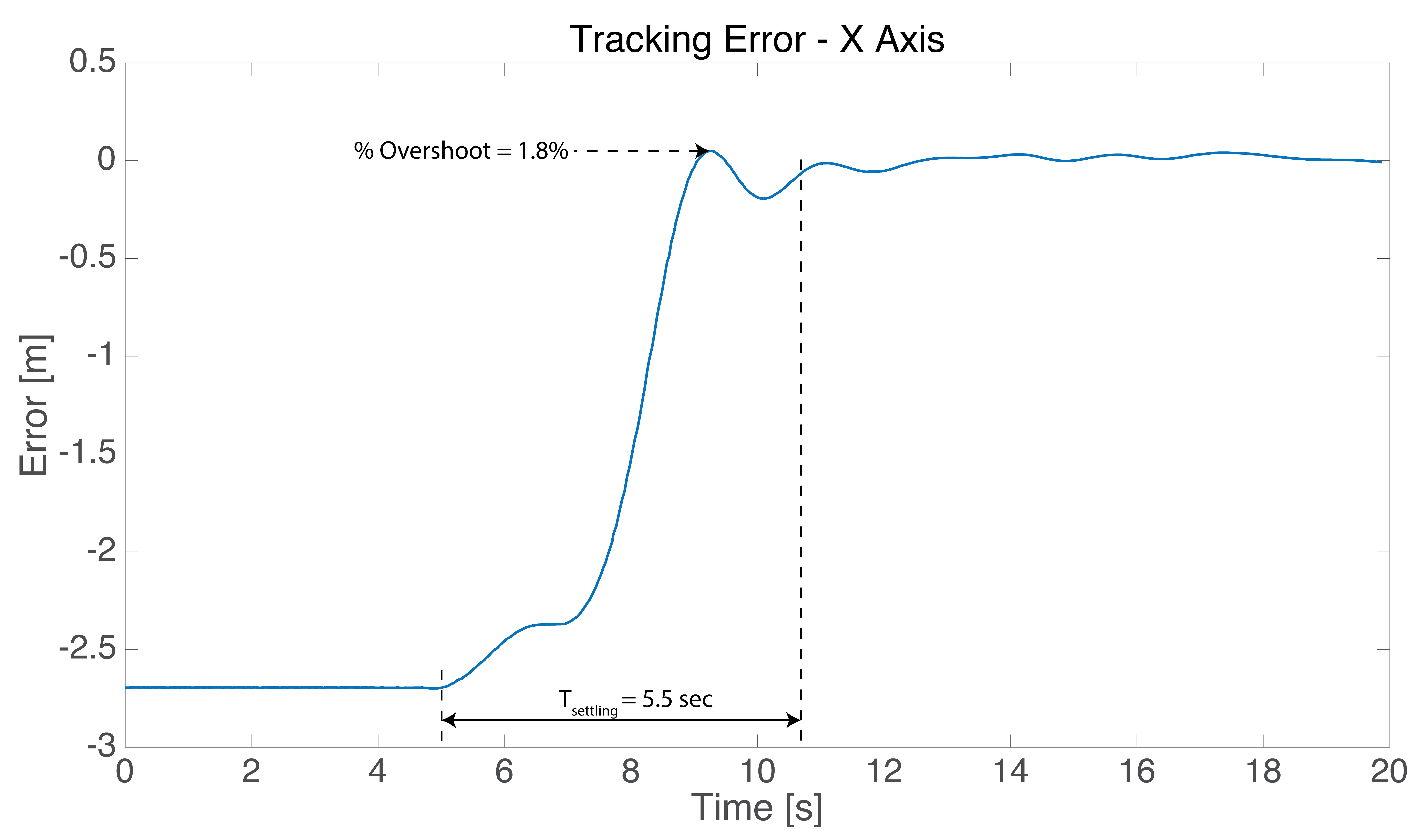}
  \caption{The error in response to a waypoint change, or a step input, results in a settling time of approximately $5.5$sec and a percent overshoot of only $1.8\%$. The X axis was chosen because the step input for this direction was the largest, at over $2.5$m, whereas the Y input is only $1$m.}
  \label{F.x_error1}
\end{figure}

To quantify the controller performance, the error in response to a waypoint change, or step input, is shown in Fig.~\ref{F.x_error1}. The X axis error is chosen as the worst case scenario in the experiment, having a step input of over $2.5$m versus only $1$m on the Y axis.

The controller's performance is quite good to step inputs, with an approximate settling time of $5.5$sec, and a percent overshoot of only $1.8\%$. A comparison between the simulated and experimental results is outlined in Table~\ref{T.exp_results}. While the experimental results do have a slightly longer settling time, and more overshoot, this is not surprising. In a real world application, the controller is subject to disturbances such as ground effects from propeller wash, since the drone is operating close to the ground and desks.

\begin{table}[htb]
\renewcommand{\arraystretch}{1.2}
\caption{Simulation vs Experimental Evaluation}
\label{T.exp_results}
\centering
\begin{tabularx}{1\columnwidth}{@{\extracolsep{\stretch{1}}}*{3}{c}@{}}

\bfseries Experiment & \bfseries Overshoot [\%]& \bfseries Settling Time [sec]\\
\hline
\hline
Simulated ePFC &  0\%  & 5 \\
Experimental ePFC &  1.8\%  &  5.5 \\

\hline
\end{tabularx}
\end{table}

As the drone approaches the obstacle, the repulsive potential pushes the drone around it as expected. In this experiment, the drone avoids the obstacle by a margin of approximately $0.5$m. Thus, this demonstrates that the drone can successfully avoid obstacles.

In addition to tracking static targets and virtual waypoints, a final test is performed in which the ARDrone is commanded to follow the target as it moves about the lab environment in an arbitrary pattern. During this experiment, the drone must maintain a safe distance at all times and should always face the target. This task was performed several times to evaluate the performance. In each of the tests the drone successfully completes the task. Even under extreme circumstances (e.g., very fast maneuvers) the drone is able to recover and maintain the desired behavior. Figure~\ref{F.dynamicTracking} shows frames from a video~\cite{youtube} taken of the drone performing this task. In the video it can clearly be seen that the drone follows the target around while always maintaining the proper heading to face the target. For more information of this test please see the video at this link: 
\url{https://www.youtube.com/watch?v=v85hs8-uc1s}

\begin{figure}[htb]
\centering
\includegraphics[width=1\columnwidth]{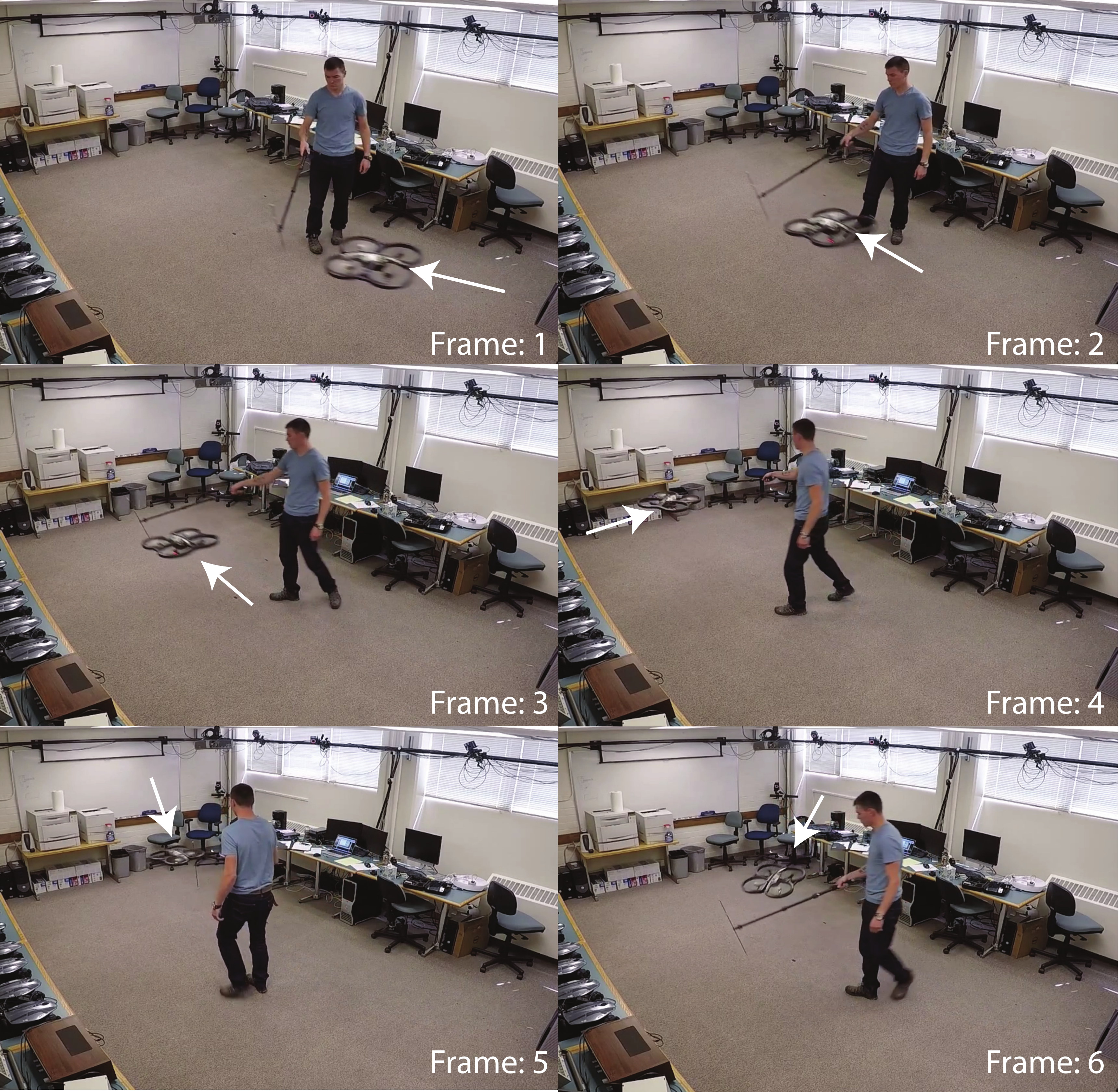}\\
\caption{The third experiment demonstrates the drone tracking a dynamic target, while maintaining the proper heading to always face the target.}
\label{F.dynamicTracking}
\end{figure}

\section{Conclusions}\label{S.conclusion}
This paper presented an extended potential field controller (ePFC) which augments the traditional PFC with the capability to use relative velocities between a drone and a target or obstacles as feedback for control. Next, the stability of the ePFC was proven using Lyapunov methods. Additionally, the presented controller was simulated and its performance relative to a tradition PFC was evaluated. The evaluation shows that the ePFC performs significantly better than a traditional PFC by reducing overshoot and settling time when navigating between waypoints. Finally, experimental results were presented which showed the actual performance of the controller.

Future work may include using an experimental system with completely on-board sensing capabilities. For a completely on-board implementation, sensing hardware would be required which would allow the drone to both localize itself in the environment as well as obstacles. Promising possibilities exist and are being utilized by various research groups (e.g, LIDAR, RGBd cameras). Potentially, the front facing camera on the ARDrone 2.0 could be used for localization using computer vision algorithms if relative position is all that is required for the application. Aside from sensing capabilities, this algorithm is very computationally inexpensive and can run on nearly any flight controller or other on-board computer, and therefore the processing requirements can be easily met with nearly any off the shelf solution today.

The proposed ePFC can be extended to a collaborative potential field control for multi-UAV for in-flight collision avoidance as well as obstacle avoidance \cite{Nguyen_IEEE_CNS, La_IROS09, La_SMC09, La_ICIEA, Nguyen_CCC} even in noisy environments where the UAV's pose is affected by localization sensors' noise \cite{La_ICRA10, Dang_MFI}.  Cooperative sensing and learning \cite{La_IEEE_CST, La_CASE13, La_SMCB_2013} for multi-UAV can be developed based on this collaborative potential field control.

\bibliographystyle{IEEEtran}
\bibliography{CASE_2016_bibliography}

\end{document}